\def\year{2021}\relax
\documentclass[letterpaper]{article} 
\usepackage{aaai21}  
\usepackage{times}  
\usepackage{helvet} 
\usepackage{courier}  
\usepackage[hyphens]{url}  
\usepackage{graphicx} 
\usepackage{natbib}  
\usepackage{caption} 
\usepackage{makecell}

\usepackage{thmtools, thm-restate}
\usepackage{subcaption}
\usepackage{bm}
\newlength{\twosubht}
\newsavebox{\twosubbox}
\usepackage{bbm}
\usepackage{array}
\usepackage{algorithm}
\usepackage{algpseudocode}

\usepackage{enumitem}
\usepackage{amsmath,amsfonts,amssymb, amsthm}
\usepackage{mathtools}

\makeatletter
\newcommand*\dashline{\rotatebox[origin=c]{90}{$\dabar@\dabar@\dabar@$}}
\makeatother

\newtheorem{proposition}{Proposition}

\newtheorem{theorem}{Theorem}

\newtheorem{assumption}{Assumption}

\newcommand{\indep}{\perp\!\!\!\perp}

\title{The Importance of Modeling Data Missingness \\ in Algorithmic Fairness: A Causal Perspective}
\author {
        Naman Goel,\textsuperscript{\rm 1}
        Alfonso Amayuelas, \textsuperscript{\rm 2}
        Amit Deshpande, \textsuperscript{\rm 3}
        Amit Sharma \textsuperscript{\rm 3}
        \\
}
\affiliations {
    \textsuperscript{\rm 1} ETH Z\"{u}rich, \textsuperscript{\rm 2} EPFL Lausanne, \textsuperscript{\rm 3} Microsoft Research \\
    naman.goel@epfl.ch, amitdesh@microsoft.com, amshar@microsoft.com
}

\begin{document}
\maketitle

\begin{abstract}
Training datasets for machine learning often have some form of missingness. For example, to learn a model for deciding whom to give a loan, the available training data includes individuals who were given a loan in the past, but not those who were not. This missingness, if ignored, nullifies any fairness guarantee of the training procedure when the model is deployed. Using causal graphs, we characterize the missingness mechanisms in different real-world scenarios. We show conditions under which various distributions, used in popular fairness algorithms, can or can not be recovered from the training data. Our theoretical results imply that many of these algorithms can not guarantee fairness in practice. Modeling missingness also helps to identify correct design principles for fair algorithms. For example, in multi-stage settings where decisions are made in multiple screening rounds, we use our framework to derive the minimal distributions required to design a fair algorithm. Our proposed algorithm decentralizes the decision-making process and still achieves similar performance to the optimal algorithm that requires centralization and non-recoverable distributions.
\end{abstract}      

\section{Introduction}
Algorithmic decision making is increasingly being used in applications of societal importance such as hiring~\cite{hiring}, university admissions~\cite{timeshigher}, lending~\cite{forbes}, predictive policing~\cite{aaas}, and criminal justice~\cite{compas}. It is well known that algorithms can learn to discriminate between individuals based on their sensitive attributes such as gender and race, even if the sensitive attribute is not explicitly used~\cite{barocas2016big}. As a result, there has been a lot of recent research on ensuring the fairness of automated~\cite{barocas2018fairness}, and machine-aided~\cite{green2019principles} decision making. A common way to mitigate fairness concerns is to include fairness constraints in the training process. For example, demographic parity constrains acceptance probability to be same across sensitive groups whereas equalized odds constrains acceptance probability given true outcome to be the same.

\begin{figure}[H]
\centering
\includegraphics[width=\columnwidth]{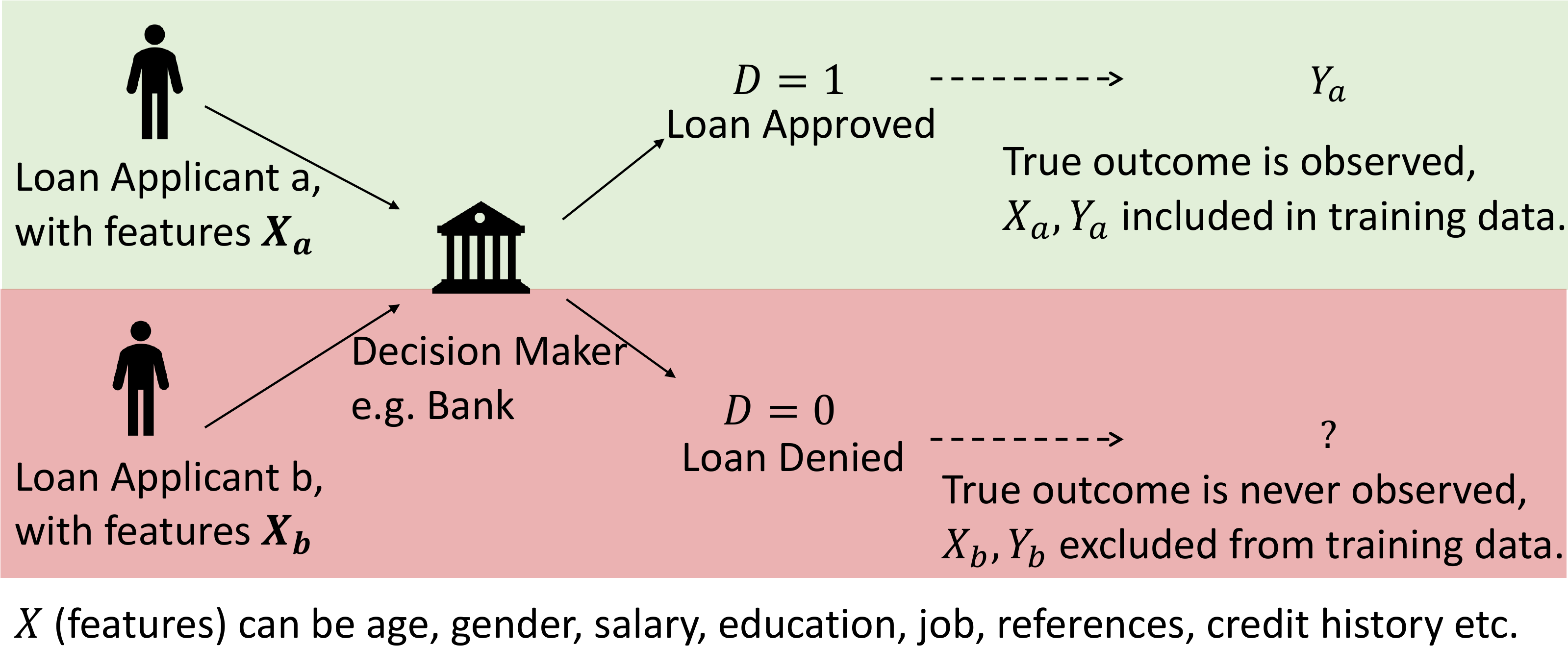}
\caption{Missingness in Training Data due to Past Decisions} 
\label{fig:intro_figure}
\end{figure}

These fairness-enhanced classifiers are generally trained and evaluated on datasets containing historical outcomes and features. However, a critical (and often unavoidable) limitation of this approach is that the datasets present only one side of the reality. There are often systematic biases that determine whose data is included in (or excluded from) the datasets. For example, consider the German credit dataset~\cite{dua2017uci} that contains profiles of people whose loans were approved and the outcome measured is whether they repaid their loan or not. The phenomenon is further illustrated in Figure~\ref{fig:intro_figure}. One may train a classifier on this data that satisfies certain fairness constraints (e.g., as done by~\citet{hardt2016equality}) to predict potential defaulters. But when the classifier is used to decide credit-worthiness of future applicants, it can be arbitrarily unfair, even if it satisfies the fairness constraints on the training data~\cite{kallus2018residual}. The reason is, in the real-world, the classifier needs to decide for all incoming applications, whereas the training dataset only contains profiles of people who were given loans. 
Similar results follow for applications in recidivism prediction~\cite{lakkaraju2017selective} and healthcare~\cite{nordling2019fairer,rajkomar2018ensuring}.

In this paper, we  provide a formal framework to reason about the effect of data missingness on algorithmic fairness, a generalized version of the special cases discussed above. We use a recently proposed causal graph framework~\cite{mohan2020graphical} to model how different types of past decisions affect data missingness. We formally show, for example, that if the past decisions are fully automated, missingness in data is relatively easier to handle. On the other hand, if missingness is caused by human (or machine-aided) decisions, then it is often impossible to learn many distributions correctly, no matter how large the dataset is. This impossibility implies that the fairness guarantees of algorithms using such distributions cannot hold in practice.

Modeling data missingness also facilitates the design of better fairness algorithms for practical use. We demonstrate this by considering a multi-stage decision making scenario. At each stage of the selection process, decision makers observe new features about the individuals who pass the previous stage, and decide whether to forward an individual to the next stage or not. We show how to model data missingness in this setting using the causal graph framework and reason about the distributions that can (and can not) be used in a fair algorithm for multi-stage setting. We use these observations to propose the first detail-free and decentralized algorithm for multi-stage settings without compromising on accuracy, unlike past work that requires centralization and knowledge of non-recoverable distributions.

\subsubsection{Summary of Our Main Contributions.}
\begin{itemize}
    \item We provide a causal graph-based framework for modeling data missingness in common scenarios from the fairness literature. We provide results on which parts of the joint data distribution can be recovered from incomplete available data and which cannot be recovered. Critically, in many  scenarios of missingness, the distributions used in common fairness algorithms are not recoverable.
    
    \item We show how the above results can guide the design of fair algorithms in practice by proposing a detail-free, decentralized and fair algorithm for multi-stage setting. Our theoretical and empirical analysis on three real-world datasets shows that the algorithm provides same utility as an optimal algorithm which assumes full centralization and knowledge of non-recoverable distributions.
\end{itemize}

\section{Background and Related Work}
Distribution shift is a well-known problem in machine learning. This shift may be, for example, a covariate ($P(X)$) shift, a concept ($P(Y|X)$) shift or even a target ($P(Y)$) shift. While this is a broad research domain, we consider the specific setting of decision-making where the data available for training has a systematic missingness due to past decisions.

\subsubsection{Notation.} We use the following notation throughout the paper. We use $X$ for the observed features about individuals, $U$ for the unobserved features, $Z$ for the sensitive attribute, $D$ for past binary decisions, $Y$ for the outcome realized under the given decision in the past, and $Y^c$ for the counterfactual outcome if the past decision was reversed. The term `unobserved variables' here means that such variables are not at all measured (not considered during data collection or are impossible to be measured) whereas the term `missingness in data' means that values of the observed variables are not present in some rows of the dataset.

\subsection{Data Missingness in Fairness}
While data missingness problem in fairness has received some attention, different papers consider different scenarios and lack a common terminology and framework for data missingness. We categorize the types of missingness studied in the fairness literature in Table~\ref{tab:related}. The first category corresponds to datasets where the outcome $Y$ is missing for certain rows and the second category addresses a bigger problem when entire rows of $(X, Y, Z)$ are missing. Thus, the first category is subsumed in the second category and addresses a relatively easier problem. \citet{KilGomSchMuaVal20} and \citet{ensign2018decision} may also be placed in the first category since they consider sequential decision making settings and there is no hard constraint that $X, Z$ are missing for $D = 0$. \citet{kallus2018residual} study a setting where $X, Y, Z$ all are missing for $D=0$ and propose a weighting-based solution that assumes access to an unlabeled dataset with no missingness. We extend their work by providing a formal graphical framework to reason about missing data  that can apply to any decision-making scenario.

The third category is for specialised scenarios where $D$ directly affects $Y$. The focus in our paper is on the second category where the decision itself does not affect a person's outcome (e.g., awarding a loan \textit{does not} increase the chances of a person paying it back), which is perhaps the most common setting for algorithmic decision-making. 
Finally, \citet{liu2018delayed, hu2018short, hu2019disparate, milli2019social,tabibian2019optimal, mouzannar2019fair, creager2019causal} consider the effect of $D$ on future $X, Z$ distribution in sequential decision making, which is a different problem but also presents missingness related challenges.

\begin{table}[ht]
\centering
\begin{tabular}{ | m{2cm} | m{2cm}| m{3cm} | }
\hline
Missingness in Variables & D affects Y? & Related Work \\ 
\hline
Only $Y$ is missing in the rows corresponding to $D = 0$. & No & \cite{lakkaraju2017selective} \\ 
\hline
Entire rows ($X, Y, Z$) corresponding to $D = 0$ are missing. & No & This Paper + \cite{kallus2018residual,ensign2018decision,KilGomSchMuaVal20} \\ 
\hline
Only $Y^c$ is not observed, $X, Y, Z$ have no missingness. & Yes & \cite{jung2018algorithmic, coston2020counterfactual, kallus2019assessing} \\ 
\hline
\end{tabular}
\caption{Categorization of Related Work}
\label{tab:related}
\end{table}

\subsection{Causal Graphs for Data Missingness}
To model data missingness in a principled manner, we will use the causal graph framework proposed by~[\cite{mohan2020graphical, mohan2013graphical, mohan2014graphical, ilya2015missing}]. The main idea in this framework is to model the missingness mechanism as a separate binary variable. If the missingness variable takes a value $0$, we observe the true value of the corresponding variable, otherwise the value is missing. Which variables affect the missingness variable decide the impact of missing data on estimating key quantities using the observed data. For example, if the missingness of a variable $X$ is caused by a variable independent from all others (e.g., a uniform random variable), then the missingness has no effect on computing $P(X)$. However, if the missingness of $X$ is caused by itself (e.g., people with $X<20$ deciding not to reveal their data), then using the d-separation criteria~\cite{pearl2009causality} and the results of~\citet{mohan2020graphical}, we can show that it is impossible to estimate (recover) $P(X)$. An illustrative example is included in the supplementary material\footnote{The supplementary material is available on the first author's website \url{https://goelnaman.github.io}.}; more details on identification with missing data are in \citet{bhattacharya2020identification,nabi2020fulllaw}.

\section{Implications of Data Missingness for Fair Machine Learning Algorithms}\label{sec:implications}
We consider the settings where past binary decisions impact missingness in the training data $(X,Y,Z)$. If the decision $D$ for an individual was $1$, then the corresponding row containing $(X,Y,Z)$ is present in the dataset otherwise it is missing. Such missingness is perhaps the most common in datasets used in the fairness literature, including in loan decision datasets~\cite{dua2017uci} (no data from declined loan applicants) or law enforcement datasets such as search for a weapon~\cite{goel2016precinct} (no data for the individuals who are not searched by police). In this section, we show how missingness affects fairness algorithms by corrupting their estimation of error rates (for \emph{equalized odds}~\cite{hardt2016equality}), allocation rates (for demographic parity~\cite{zafar2017fairness}), and probabilities $P(Y|X)$, $P(Y|X,Z)$, $P(X,Z)$ and $P(X)$. All proofs are in the supplement.

\subsection{Algorithms Requiring Error Rate or Allocation Rate Estimates}
\begin{figure}[H]
\centering
\includegraphics[width=3cm]{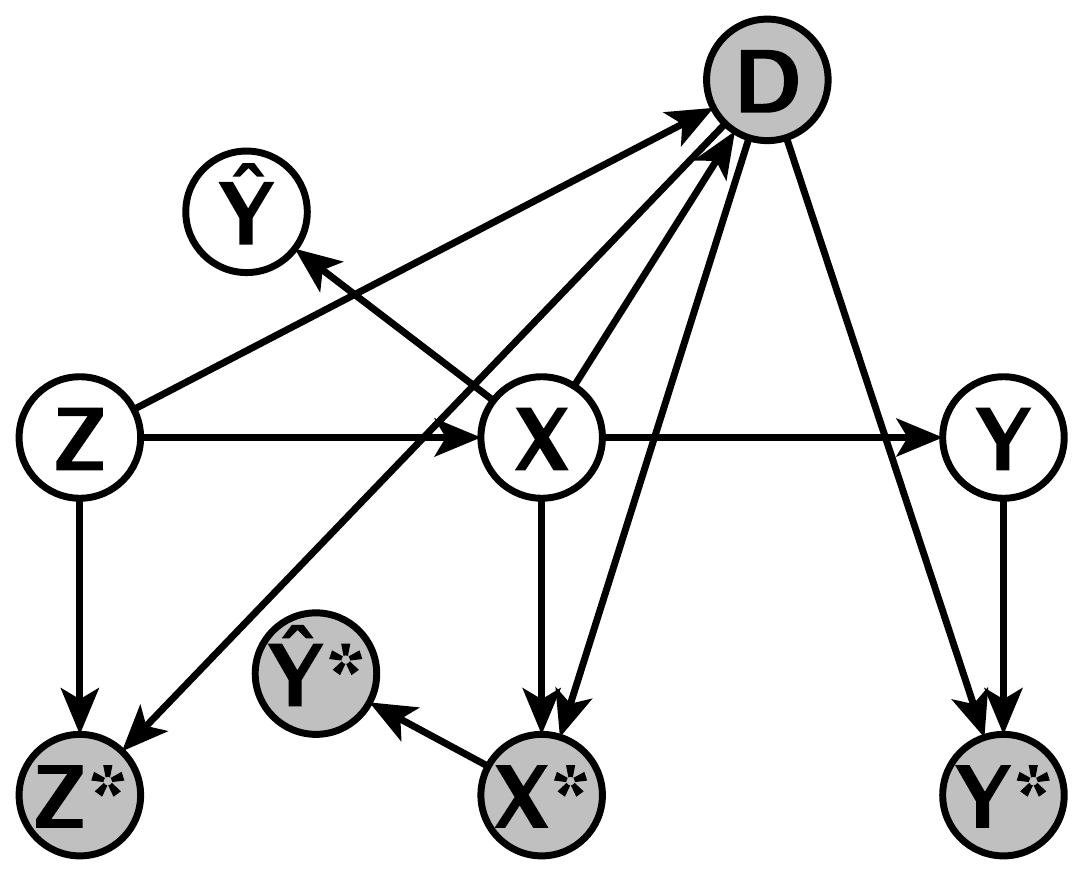}
\caption{Data Missingness Mechanism as Causal Graph. \textit{Shaded nodes are observed, unshaded ones are not observed in the training data. For instance, the node $Y^*$ denotes the outcomes for people whose data is observed and $Y$ denotes the (possibly unobserved) outcomes for everyone.}}

\label{fig:error-rate}
\end{figure}

\begin{figure*}[htp]
\sbox\twosubbox{%
  \resizebox{0.98\textwidth}{!}{%
    \includegraphics[height=0.1cm]{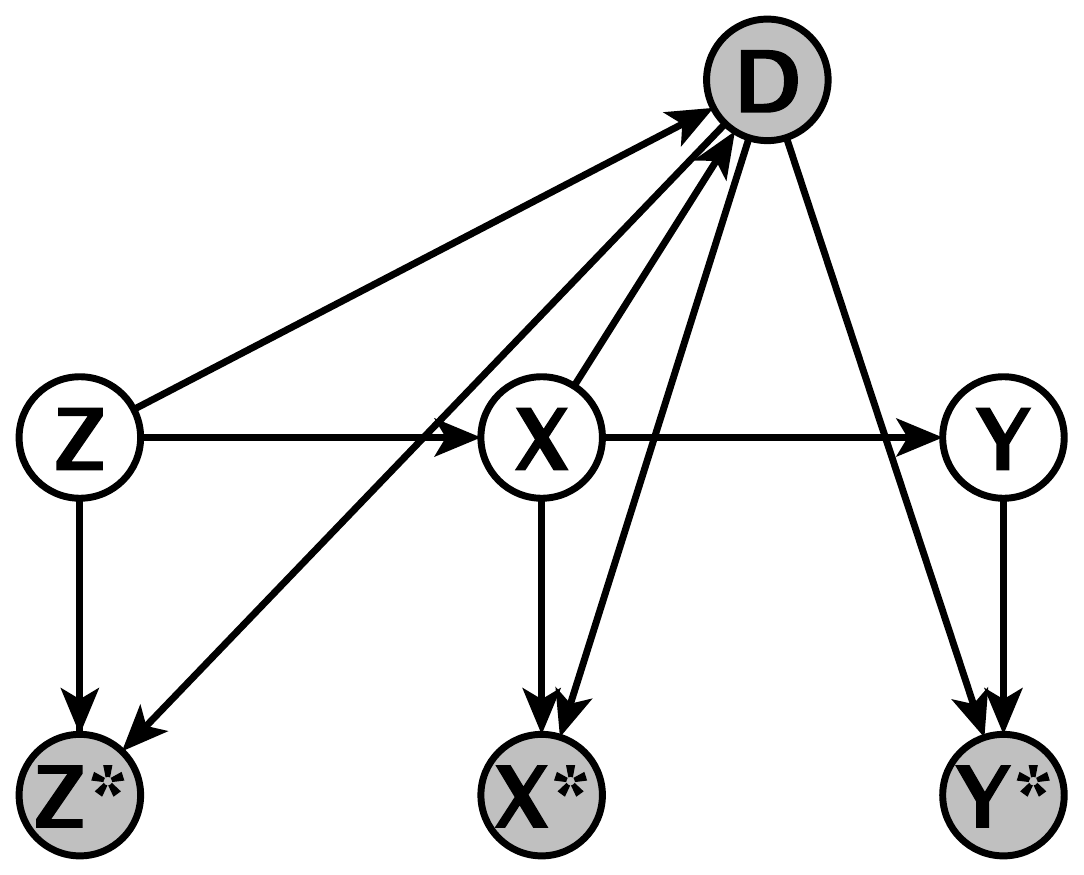}%
    \includegraphics[height=0.1cm]{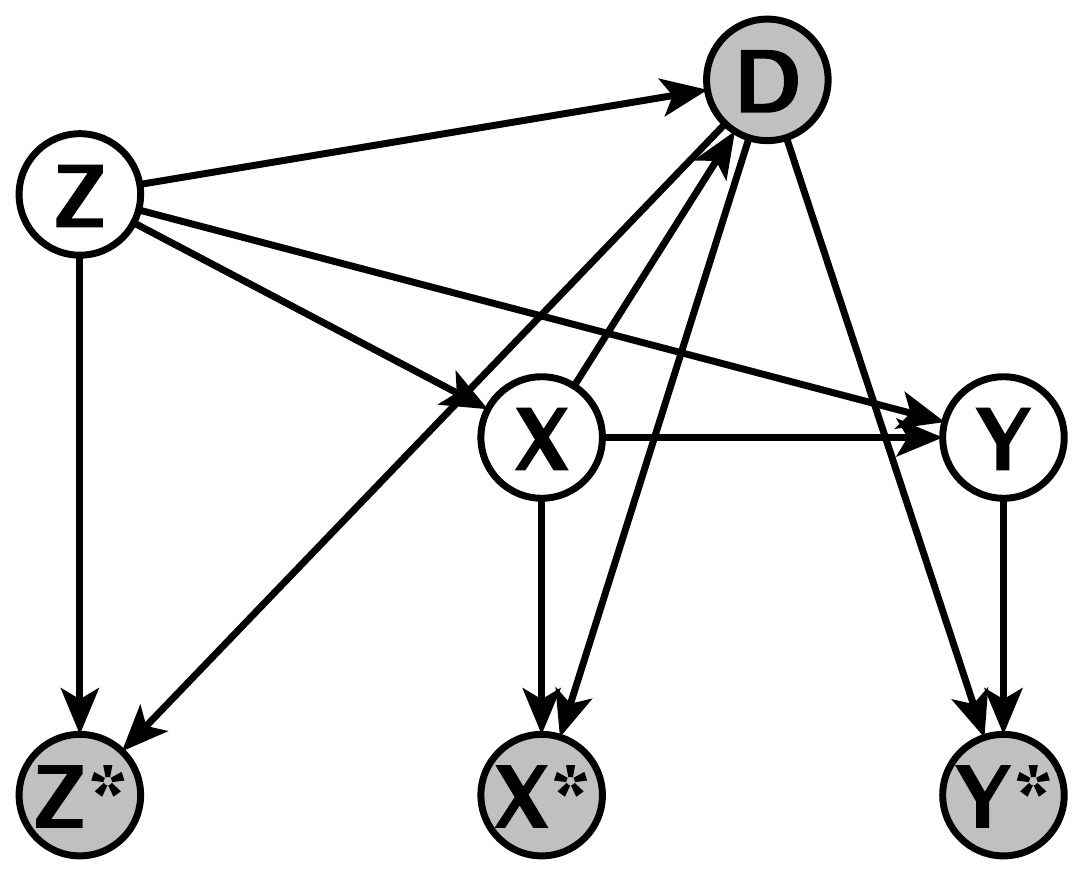}%
    \includegraphics[height=0.1cm]{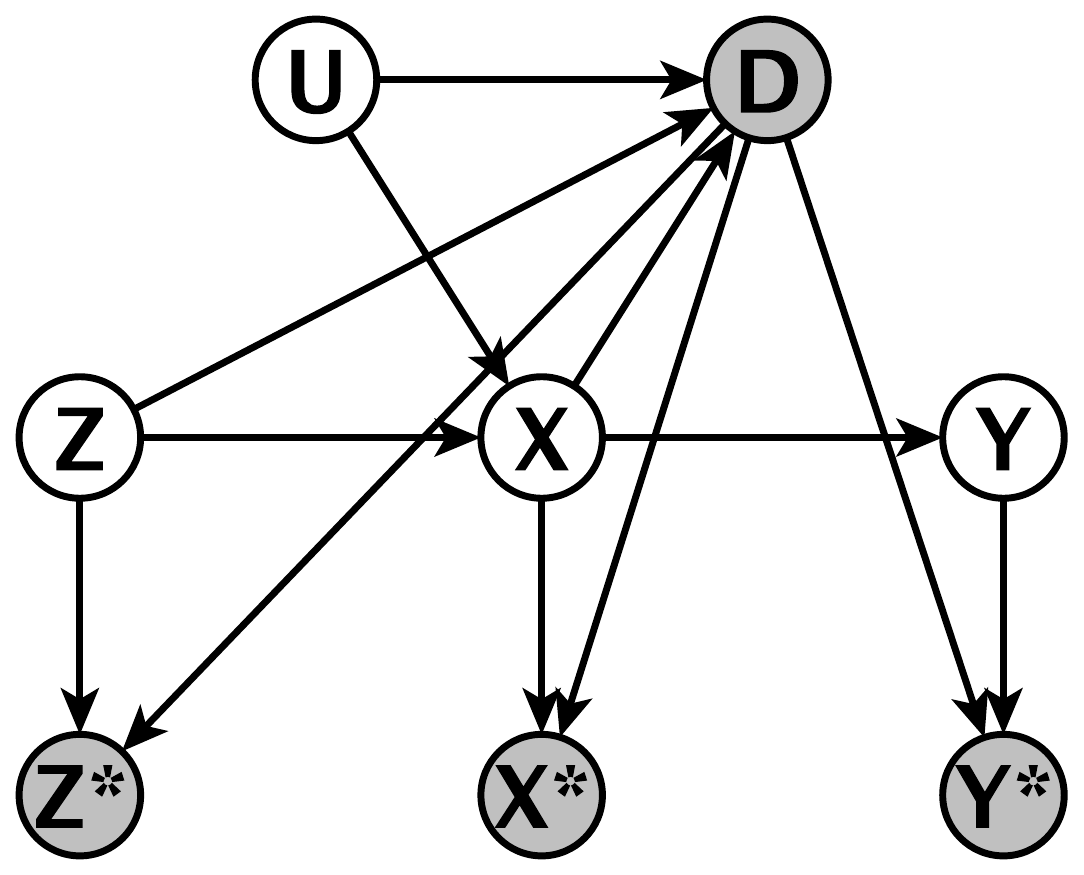}%
    \includegraphics[height=0.1cm]{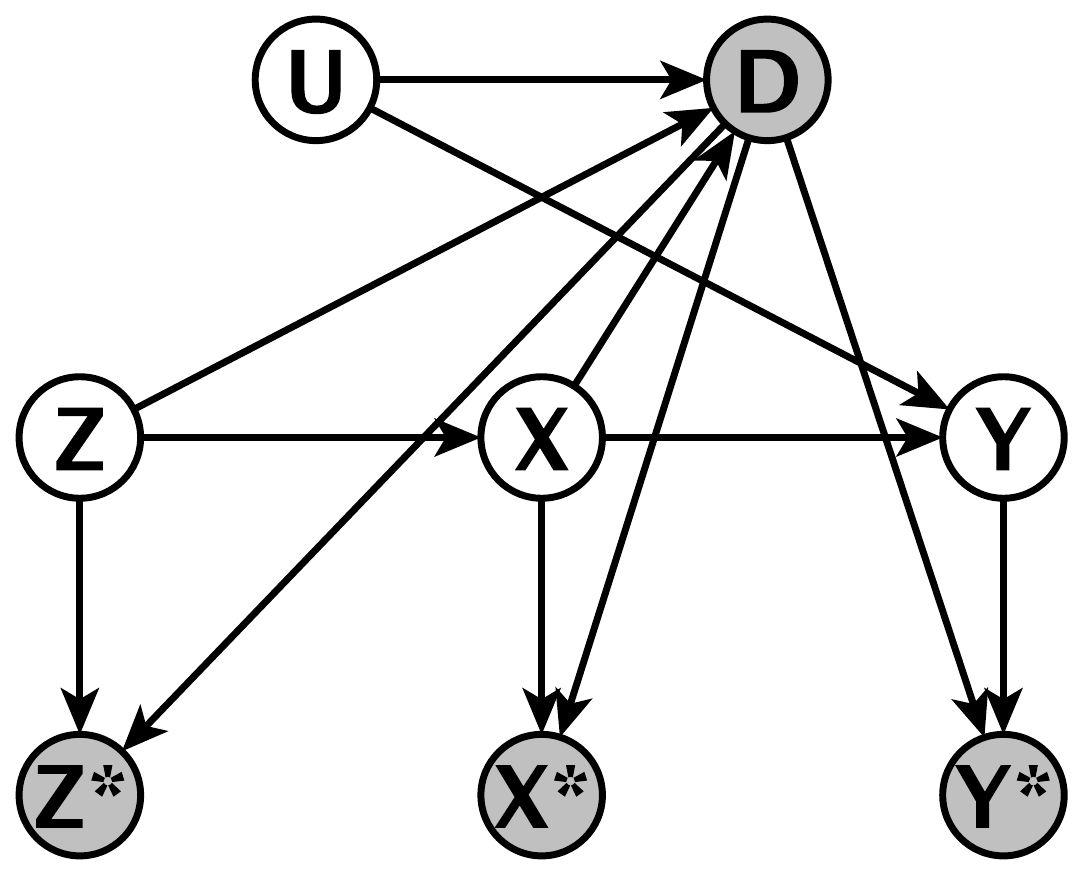}%
    \includegraphics[height=0.1cm]{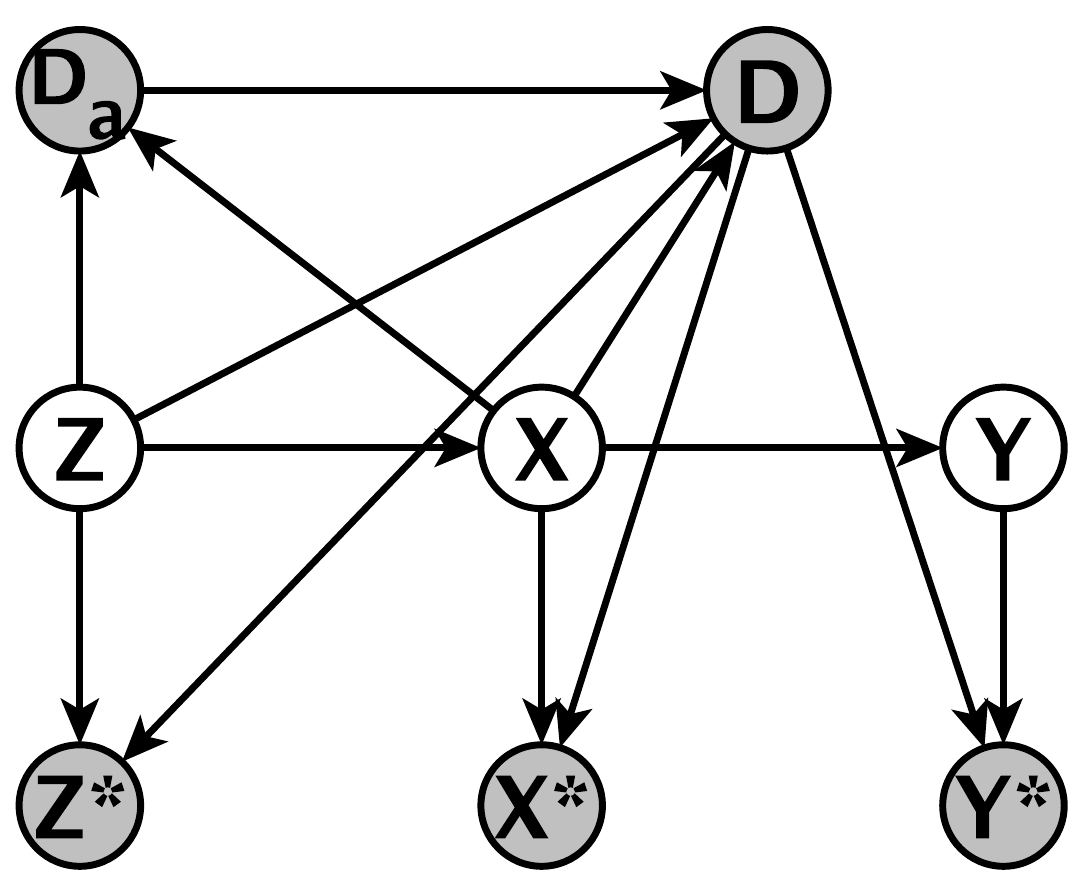}%
    \includegraphics[height=0.1cm]{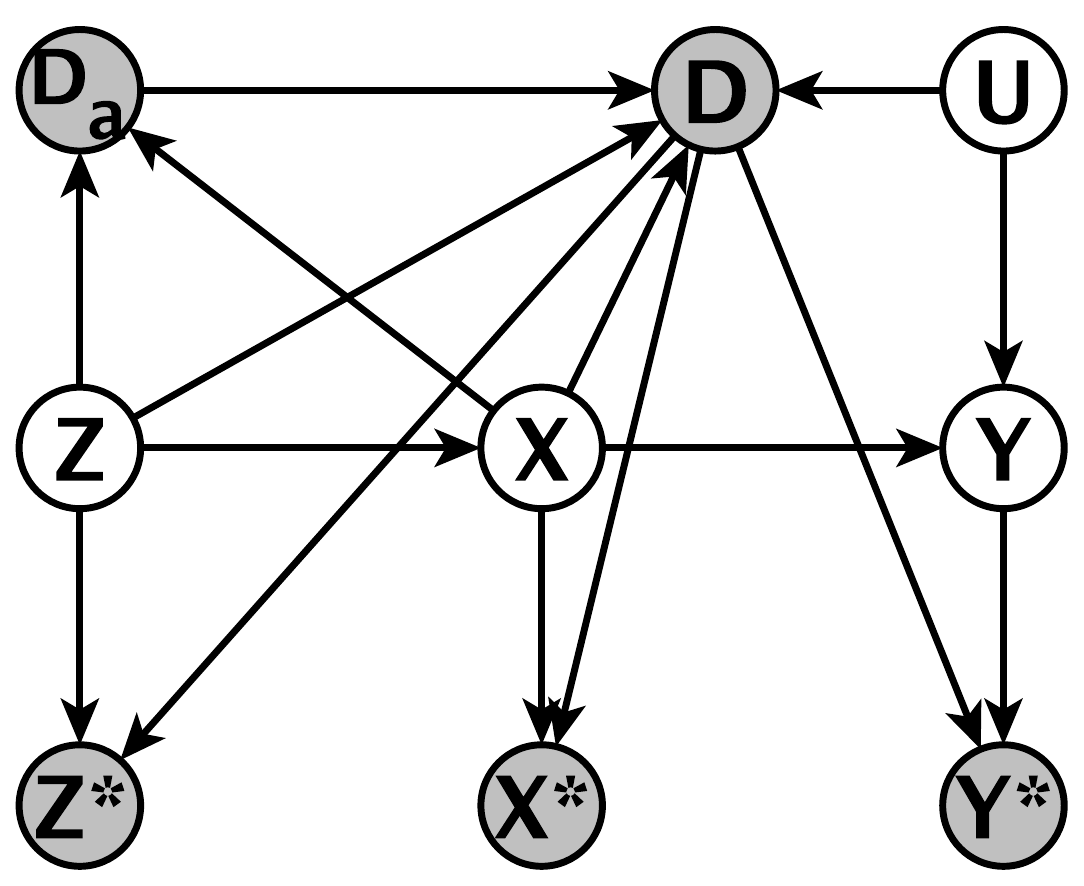}%
  }%
}
\setlength{\twosubht}{\ht\twosubbox}


\centering

\subcaptionbox{\label{i}}{%
  \includegraphics[height=\twosubht]{case2.pdf}%
}%
\subcaptionbox{\label{ii}}{%
  \includegraphics[height=\twosubht]{case4.pdf}%
}%
\subcaptionbox{\label{iii}}{%
  \includegraphics[height=\twosubht]{case3_z.pdf}%
}%
\subcaptionbox{\label{iv}}{%
  \includegraphics[height=\twosubht]{example8.pdf}%
}
\subcaptionbox{\label{v}}{%
  \includegraphics[height=\twosubht]{case7.pdf}%
}
\subcaptionbox{\label{vi}}{%
  \includegraphics[height=\twosubht]{case8_z.pdf}%
}%
\caption{Data Missingness due to Fully Automated (\ref{i},\ref{ii}), Human (\ref{iii}, \ref{iv}) and Machine-Aided Decision Making (\ref{v},\ref{vi}).}
\label{fig:individual_case}
\end{figure*}

Consider the causal graph shown in Figure~\ref{fig:error-rate}. In this graph, nodes $X$, $Z$ and $Y$ represent the random variables for non-sensitive features, the sensitive attribute and the outcome respectively. For each of these nodes, we show a starred and shaded node. These nodes represent the random variables that we actually observe in the training data. Each starred node also has an incoming arrow from a ``missingness variable". In our case, this missingness variable is the same as past decisions variable $D$. $X^* = X$, $Z^* = Z$ and $Y^* = Y$ when $D = 1$ and are missing otherwise. The incoming arrows to the missingness variable show which other variables affect the missingness in data. Figure~\ref{fig:error-rate} can be interpreted as showing the missingness mechanism for the \emph{lending} scenario as follows: the past loan decision $D$ for an individual is based on their features $X$ (and possibly $Z$) and the features $X$ determine their payback outcome $Y$. Values for $X$, $Y$, $Z$ are observed in the training data only if $D=1$ as shown by the arrow from $D$ to $X^*$, $Y^*$, and $Z^*$. This causal graph captures simplistic settings but as we will show later, it is possible to represent more complex settings in a similar way. For our current discussion, this graph is sufficient.

\subsubsection{Error Rate.} To ensure equal opportunity while predicting outcome $Y$, multiple algorithms have been proposed based on adjusting the error rates~\cite{hardt2016equality,pleiss2017fairness} of classifiers. To obtain the error rate estimate of a classifier, the standard procedure is to use the classifier to predict $\hat{Y}$ on i.i.d. samples of the data and compare it to $Y$. However, since the data is incomplete, we do not have access to $\hat{Y}$ and $Y$. Instead, we observe $\hat{Y^*}$ (predictions on the available samples) and we compare them to $Y^*$ (outcomes for the available samples). Therefore, while the true error rate of the classifier for a group $Z$ is $P(\hat{Y} | Y, Z)$, we end up estimating $P(\hat{Y}^*| Y^*, Z^*)$ due to incomplete data. Since $P(\hat{Y}^*| Y^*, Z^*) = P(\hat{Y}| Y, Z, D = 1)$ (by definition),  $P(\hat{Y}^*| Y^*, Z^*)$ estimated from the available data is not equal to  the true $P(\hat{Y}| Y, Z)$ unless $\hat{Y}$ and $D$ were independent given $Y$ and $Z$.  Using d-separation, we can confirm in Figure~\ref{fig:error-rate} that $\hat{Y} \not \indep D|Y, Z$. This leads to the following result.

\begin{proposition}\label{prop:error-rate}
For a classifier, its group error rates estimated naively from the incomplete data (with missingness mechanism shown in Figure~\ref{fig:error-rate}) are inconsistent.
\end{proposition}

\noindent This result is neither surprising nor technically challenging to obtain but its implications are often ignored while designing fairness
algorithms. The proposition implies that 1) no matter how many data samples we may have, naive error rate estimates based on data with systematic missingness will be incorrect, and 2) fairness algorithms that rely on estimating such estimates from incomplete data will fail to meet the constraints in practice. That is, if such a loan decision algorithm was deployed in practice and exposed to \emph{all} applicants, fairness would not be guaranteed.

\subsubsection{Allocation Rate.}
A follow up question is whether fairness algorithms that guarantee demographic parity also fail in practice if trained on incomplete data. For demographic parity, one doesn't need to equalize error rates across groups but only the allocation rates $P(\hat{Y}|Z)$. The question can be easily answered using a similar reasoning with the causal graph shown in Figure~\ref{fig:error-rate}. The observed estimate of allocation rate is $P(\hat{Y^*}|Z^*)= P(\hat{Y}|Z, D=1)$ whereas the true allocation rate is $P(\hat{Y}|Z)$. It is easy to see in Figure~\ref{fig:error-rate} that $P(\hat{Y}| Z, D = 1) \neq P(\hat{Y}| Z)$ since $\hat{Y}$ is not \emph{d-separated} from D given Z ($\hat{Y} \not \indep D|Z$). We can conclude that if a fairness algorithm naively estimates allocation rates from incomplete data to enforce demographic parity constraints, then it will fail to meet the constraints on the general population.

\smallskip
We next show that causal graph based modeling is a powerful framework that allows us to compute general identification results by modeling data missingness in a wide variety of scenarios, including fully automated, human and machine-assisted decision-making. It is interesting to note that in our discussion, we make no assumption about the fairness or unfairness of the past decisions that cause missingness in the training data.

\subsection{Algorithms Requiring Estimates of $P(Y|X)$, $P(Y|X,Z)$, $P(X), P(X,Z)$}
Several algorithms in the fairness literature require distributions like $P(Y|X)$, $P(Y|X, Z)$, $P(X)$ or $P(X,Z)$~\cite{celis2019classification, corbett2017algorithmic, valera2018enhancing}. However, it is unclear under what conditions these  distributions can be recovered from the available data. We consider the different ways by which past decisions are typically made: by an algorithm, by a human, or by a human based on an algorithm's recommendation. For each setting, we model the causal process by which the data may be missing and show how these variations effect the recoverability of true distributions from available data.

\subsubsection{Fully Automated Decision-Making.}
In fully automated decision-making, decisions are based on observed features. While a lending dataset may exclude people whose loans were not approved, for the included persons it is likely to include all features that were used by the loan approval algorithm to make its decision. Example causal graphs for this type of missingness are shown in Figures~\ref{i} and~\ref{ii}.

\begin{proposition}\label{prop:prop1}
For the missingness mechanism shown in Figure~\ref{i}, both $P(Y|X)$ and $P(Y|X,Z)$ can be consistently estimated from the incomplete data whereas $P(X)$ and $P(X,Z)$ are not recoverable.
\end{proposition}

\noindent The above proposition implies that, given enough data samples, the conditional distributions can be recovered easily from incomplete data even if past decisions were based on sensitive attributes. The main idea in the proof remains the same as we discussed for Proposition~\ref{prop:error-rate} i.e. using the d-separation criteria to determine whether $Y$ is conditionally independent of the missingness variable $D$ given $X$ or given $X$ and $Z$. However, the observation in this Proposition changes as soon as we relax one assumption and consider the causal graph shown in Figure~\ref{ii} where the sensitive attribute $Z$ directly affects $Y$ (e.g., if person is likely to face direct discrimination based on $Z$ after being awarded the loan or after being hired, affecting their observed outcome).

\begin{proposition}\label{prop:prop2}
For the missingness mechanism shown in Figure~\ref{ii},  $P(Y|X, Z)$ can be consistently estimated from the incomplete data but a naive estimate of $P(Y|X)$ is not consistent. $P(X)$ and $P(X, Z)$ are not recoverable.
\end{proposition}

\noindent The above proposition says that no matter how many data samples we have, our naive estimate of $P(Y|X)$ will not converge to the true value. However, it only establishes the inconsistency of a naive estimate of $P(Y|X)$ and it remains an interesting open question if $P(Y|X)$ is not recoverable (i.e. there exists no consistent estimator). Note that in both Propositions~\ref{prop:prop1} and \ref{prop:prop2}, $P(X)$ and $P(X, Z)$ are not recoverable i.e. there exist no estimators for $P(X)$ and $P(X,Z)$ that converge to the true values, even with infinitely many data samples. Non-recoverability of a distribution is a stronger result compared to non-consistency of its naive estimator. In general, if there is a direct edge between a variable and its missingness mechanism (in this case, between $X$ and $D$), then the corresponding distribution is non-recoverable~\cite{mohan2020graphical}. Therefore, in typical datasets created from fully automated decision-making, it is not possible to ``fix" a fairness algorithm that relies on $P(X)$ or $P(X,Z)$ by modifying the estimator. The only way is to assume external knowledge as~\citet{kallus2018residual} do or collect additional data for estimating $P(X)$. In further discussion, we will only focus on $P(Y|X, Z)$, $P(X)$ and $P(X,Z)$ and will assume $Z$ does not directly affect $Y$.

\begin{table*}[ht]
\centering
\begin{tabular}{ | p{1.2cm} | p{2.3cm} | p{2.3cm} | p{2.3cm} | p{4.2cm} | p{2.4cm} |}
      \hline
       & \cite{pleiss2017fairness} with FPR Constraints & \cite{pleiss2017fairness} with FNR Constraints &  (Kamiran et al. 2012) with SP Constraints & \makecell{\\(Kamiran and Calders 2012)} & \cite{celis2019classification} with FDR Constraints ($\tau_{fdr} = 0.8$)\\
      \hline
      ADULT & \makecell{\textit{FPRD:} \\ $0.0725$ \\ ($-0.00724$)}  & \makecell{\textit{FNRD:} \\ $0.0377$ \\ ($0.00295$)} & \makecell{\textit{SPD:} \\  $-0.1137$ \\ ($-0.0390$)} & \makecell{\textit{SPD}: $-0.2410$ ($-0.0705$) \\ \textit{DI}: $0.5566$ ($0.7785$) \\ \textit{AOD}: $-0.1486$ ($0.0188$) \\ \textit{EOD}: $-0.1327$ ($0.0293$)} & \makecell{$\gamma_{fdr}$\\ $0.68$ \\ ($0.63$)} \\
      \hline
      COMPAS & \makecell{\textit{FPRD:} \\$0.061$ \\ ($-0.00155$)} &  \makecell{\textit{FNRD:} \\ $0.099$ \\ ($0.0056$)} & \makecell{\textit{SPD:} \\ $-0.2651$ \\ ($0.0229$)} & \makecell{\textit{SPD}: $-0.2340$ ($-0.1188$ )\\ \textit{DI}: $0.7402$ ($0.8342$) \\ \textit{AOD}: $-0.2237$ ($-0.0946$) \\ \textit{EOD}: $-0.2266$ ($0.0111$ )} & \makecell{N/A}\\
      \hline
\end{tabular}
\caption{Difference in Test (and Train) Performance of Fair ML Algorithms under Training Data with Missingness. \textit{FPR stands for false positive rate, FNR for false negative rate, FDR for false discovery rate, SP for statistical parity, AO for average odds, EO for equal opportunity, DI for disparate impact, and $\gamma_{fdr}$ is the meta-fair classifier's fairness metric (closer to $1$ is better for DI and $\gamma_{fdr}$).  The suffix D denotes the difference in corresponding metric between the privileged and the unprivileged groups.}}
\label{tab:synthetic}
\end{table*}

\subsubsection{Human Decision-Making.}
A distinguishing characteristic of human decision-making is that it may involve unobserved features. \citet{lakkaraju2017selective} provide an example: a human judge can observe whether a defendant is accompanied by their family during trial and this may also affect their outcome (whether they recidivate or not). In the resulting training data, however, this feature may be absent (therefore, called unobserved). In addition, the recidivism outcome and features would only be available for defendants that were released. To illustrate, consider the causal graphs shown in Figures~\ref{iii} and \ref{iv} (obtained from Figure~\ref{i} by adding $U$ for unobserved features). The difference between Figures~\ref{iii} and \ref{iv} is whether $U$ affects $Y$ directly or through $X$.

\begin{proposition}\label{prop:non-recov-unobserved}
For the missingness mechanism shown in Figure~\ref{iii}, $P(Y|X, Z)$ can be consistently estimated whereas for the mechanism shown in Figure~\ref{iv}, $P(Y|X, Z)$ is not recoverable. $P(X)$ and $P(X, Z)$ are not recoverable in either of the two mechanisms.
\end{proposition}

\subsubsection{Machine Aided Decision-Making.}
Another common form of decision making is machine aided decision making or algorithm-in-the-loop decision making. In this case, a human receives feedback from an algorithm before making their decision. Consider the causal graphs shown in Figures~\ref{v} and \ref{vi} (obtained from previous graphs by adding a variable $D_a$ representing algorithm's feedback).

\begin{proposition}
For the missingness mechanism shown in Figure~\ref{v}, $P(Y|X, Z)$ can be consistently estimated from the incomplete data whereas for the mechanism shown in Figure~\ref{vi},  $P(Y|X, Z)$ is non-recoverable. $P(X)$ and $P(X,Z)$ are not recoverable in both these cases.
\end{proposition}

\noindent The above proposition implies that the new variable $D_a$ in hybrid decision making doesn't pose additional challenges in the estimation of  $P(Y|X, Z)$ distributions but the usual challenges due to human involvement continue to exist.

\subsection{Empirical Implications of Data Missingness}
\citet{kallus2018residual} showed that a supposedly equal opportunity classifier of~\citet{hardt2016equality} trained on New York Stop, Question and Frisk dataset~\cite{sqf-dataset} does not ensure equal opportunity in the general NYC population: it wrongly targets up to 20\% of white-Hispanic, 16\% of other, and 14-15\% of black innocents, but only 11\% of white-non-Hispanic innocents. We present several semi-synthetic experiments using the Adult~\cite{dua2017uci} and the COMPAS~\cite{compas-dataset} datasets to show that this is a more general phenomenon, across different datasets, algorithms and fairness metrics. 

We created synthetic missingness in these datasets based on the following procedure. We trained a logistic regression classifier from the scikit-learn library~\cite{pedregosa2011scikit} and then deleted the records from the training set for which the classifier's predicted probability for the favourable class was below a certain threshold. We chose the threshold to be low enough ($0.06$ for Adult and $0.55$ for COMPAS) to ensure that the label distribution in the resulting dataset does not become very skewed. We did not delete any record from the test set (preserving the general distribution without any missingness). Note that such semi-synthetic experiments are necessary because of the unavailability of datasets with and without missingess. We then trained fairness algorithms with different underlying principles (pre-processing, in-processing and post-processing) on this semi-synthetic training data, using example code provided in IBM's AI Fairness 360 library~\cite{bellamy2019ai}. In particular, we studied the calibrated equalized odds post-processing algorithm of~\citet{pleiss2017fairness}, the reject option classifier of ~\citet{kamiran2012decision}, the re-weighting pre-processing algorithm of ~\citet{kamiran2012data}, and the meta-fair classifier of~\citet{celis2019classification}. The choice of fairness metrics  and datasets in the experiments is governed by the example code provided for respective algorithms. For the meta-fair classifier, we set the $\tau_{fdr}$ hyper-parameter to $0.8$ as provided in the example code. 

The results are summarized in Table~\ref{tab:synthetic}. For every algorithm, we compare the  fairness measures observed on the test set and the censored training set (in parenthesis). For the pre-processing algorithm by \citet{kamiran2012data}, we show the fairness measures on the test set when there is synthetic missingness in the training set, and in parenthesis, we show the fairness measures on the test set when there is no missingness in the training set. The table shows that all algorithms show significant disparities in their fairness metric due to missing data. In many cases, there is an order of magnitude difference (\textit{over 10X}) between the fairness metric on the test and (censored) train data,  validating our theoretical claims about the implications of data missingness.

\subsection{Discussion}
\subsubsection{Importance of Knowing Precise Causal Structure.}
While the causal graphs we discussed are not exhaustive, the framework (and the same proof technique) can be used for deriving recoverability results in any new scenario. We also note that it is not enough to only model the existence of relation between variables; the direction of the causal relation is also important. For example, throughout our discussion, we assumed that the features $X$ cause the outcome $Y$. However, in some settings, the reverse may be true. That is, the target variable $Y$ (e.g. skill $Y$) of a person causes/determines their features $X$ (e.g. test scores). This may affect the conclusions about estimation of distributions. Consider, for example, the graph shown in Figure~\ref{iii}. 
Reversing the edge between $X$ and $Y$ in this graph affects the corresponding observation stated in Proposition~\ref{prop:non-recov-unobserved}: a naive estimate of the distribution $P(Y|X,Z)$ is not consistent in the new graph. Thus, it is important to understand the domain of interest and reason about the effects of the data censoring mechanism on the learning algorithms accordingly.

\subsubsection{Better Data Collection Practices.} While missingness in the labels is unavoidable, missingness in the unlabeled data can be avoided by adopting better data collection practices. This may be exploited to provide better fairness guarantees, as in ~\citet{kallus2018residual}. Similarly, challenges due to human involvement can be addressed by designing data (features) collection process that better capture human decision-making, or by exploiting random variations that remove the effect of unobserved variables \cite{kleinberg2017judges}.

\section{Application: Multi-Stage Decision Making}
The framework and the results presented in the previous section are not only useful for critically evaluating existing fairness algorithms but also for designing new algorithms. In this section, we present one such example: the case of multi-stage decision-making. Multi-stage processes are common, e.g., in hiring, university admissions and even in lending. At each stage of the selection process, decision makers request or collect new features about the individuals and make decisions on whether to forward an individual to the next stage or not. Each stage of the selection process narrows down (subject to budget constraints) the pool of individuals and more features are observed in the subsequent stages for individuals who pass the previous stage.
\begin{figure}[htp]
\sbox\twosubbox{%
  \resizebox{0.5\columnwidth}{!}{%
    \includegraphics[height=0.1cm]{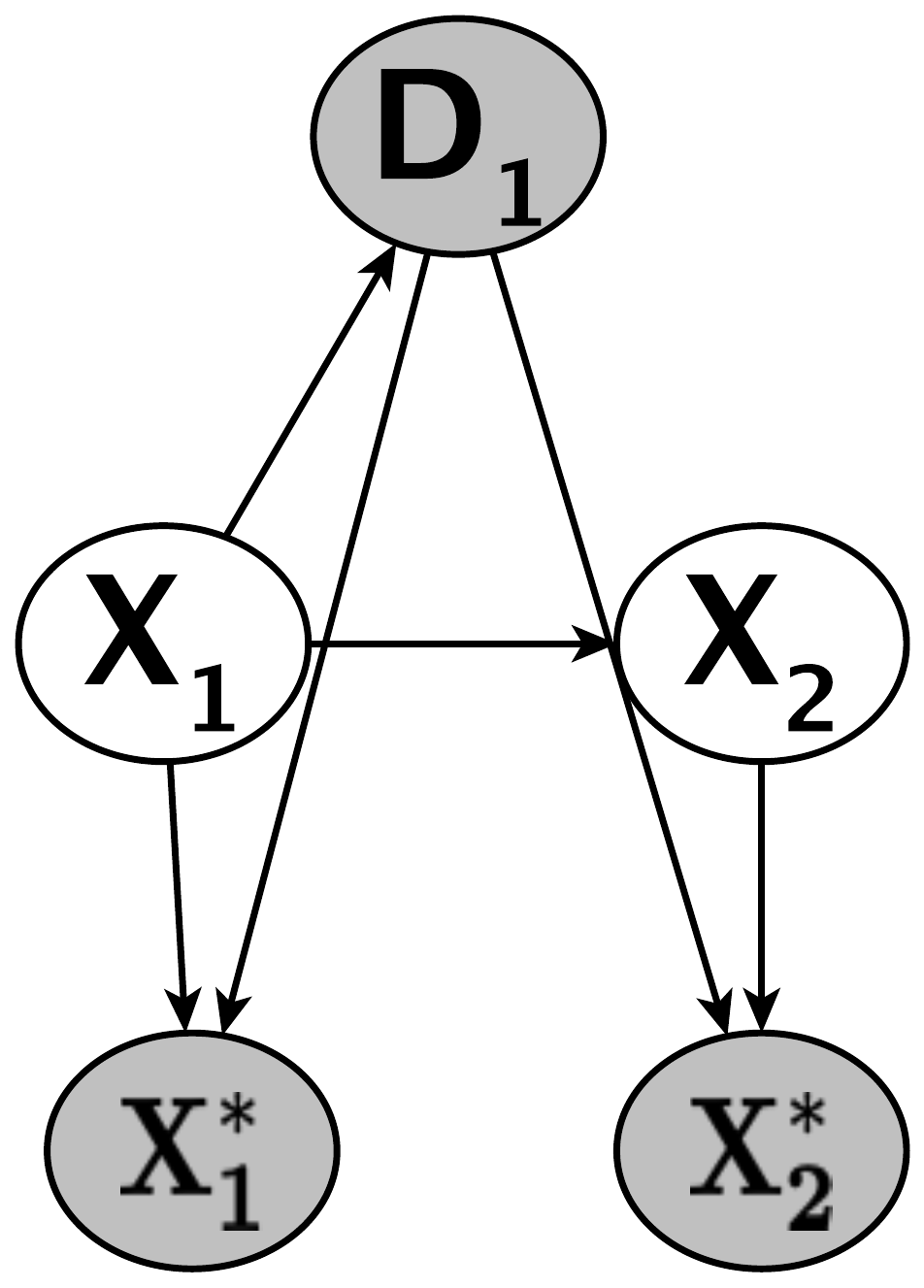}%
    \includegraphics[height=0.1cm]{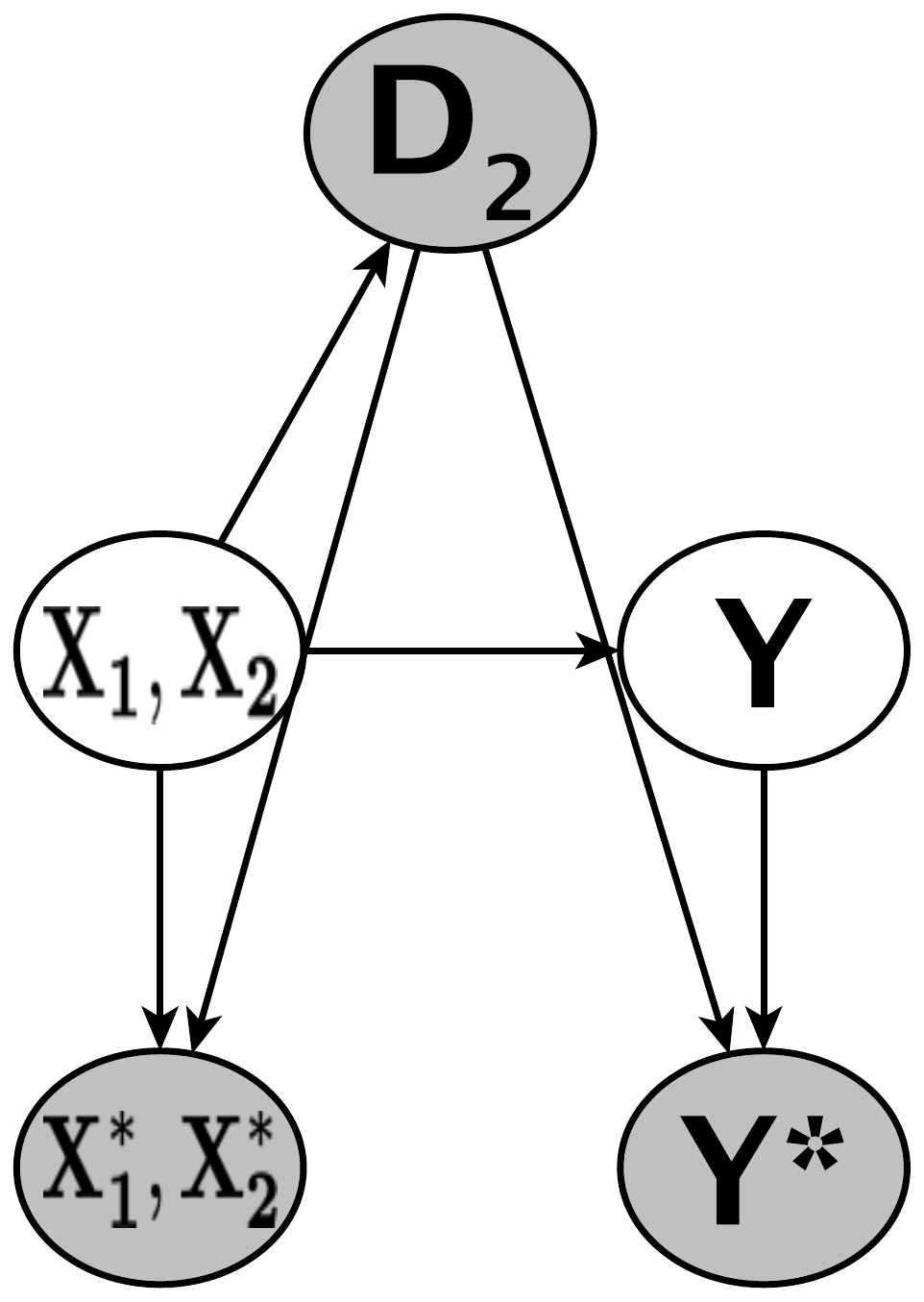}%
  }%
}
\setlength{\twosubht}{\ht\twosubbox}
\centering
\subcaptionbox{Stage 1\label{2si}}{%
  \includegraphics[height=\twosubht]{stage1.pdf}%
}%
\qquad
\subcaptionbox{Stage 2\label{2sii}}{%
  \includegraphics[height=\twosubht]{stage2.pdf}%
}%
\caption{Data Missingness in 2-Stage Process ($Z \in X_1$)}
\label{fig:2stage-graphs}
\end{figure}

\noindent Figure~\ref{fig:2stage-graphs} shows an example of a 2-stage decision process for hiring where stages represent independent organizations or teams evaluating the applicants (e.g., a recruitment agency in the first stage and the hiring entity itself in the second stage). As before, the decision $D_1$ in the first stage causes missingness in the features $X_1$ and the first stage's outcome (whether additional features $X_2$ are collected for the individual). Similarly, in the second stage, the decision $D_2$ selects the final individuals and affects missingness of both $(X_1, X_2)$ and the outcome $Y$: success outcome $Y$ is observed only for individuals who passed the second stage (and consequently, also the first stage). Typically, $(X_1, X_2)$ is also recorded only for those individuals.
 
 \begin{proposition}\label{prop:2stage}
 For the 2-stage missingness mechanism shown in Figure~\ref{fig:2stage-graphs}, $P(Y|X_1)$ and $P(Y|X_1, X_2)$ can be recovered from the incomplete data but joint distribution $P(X_1, X_2)$ is not recoverable.
 \end{proposition}

Consider a scenario where an incomplete dataset (resulting from past multi-stage decision making as shown in Figure~\ref{fig:2stage-graphs}) is available for learning distributions that we can use for a future application (which again is a similar multi-stage decision making application). \citet{emelianov2019price} propose an algorithm for a multi-stage decision-making application. However, the algorithm assumes full knowledge about the joint distribution of all features of the population, which according to Proposition~\ref{prop:2stage} is not recoverable from incomplete data. Moreover, in decentralized settings each stage makes decisions on its own and feature distribution of all the applicants may not be shared between stages. Therefore, we explore another dimension of the solution space, which is to design algorithms without requiring distributions that cannot be recovered from incomplete data.

\subsection{The $\bm{DF^2}$ Algorithm for \underline{D}etail-\underline{F}ree, \underline{D}ecentralized and \underline{F}air Multi-Stage Decision Making}
\subsubsection{Problem Formulation.} Given a pool/batch of individuals, we want to find optimal decision rules $D_i$ for each stage $i$ such that the precision of the decision rule is maximized subject to constraints on the budget (fraction of individuals to be selected at that stage) and fairness. We assume that additional features are observed at each stage, so stage 1 observes $X_1$ features, and stage $i$ observes $\{X_1, X_2, ...X_i\}$ features. \citet{emelianov2019price} have shown that when the budget is fixed, maximizing precision is equivalent to maximizing true positive rate, true negative rate, accuracy and f1-score; and minimizing false positive rate and false negative rate. Let $Y$ represent the true outcome (i.e. success in job) and $\hat{Y_i}$ denote the action taken for an individual at stage $i$ (whether the individual is given a job or not). $D_i$ is the optimization variable at stage $i$. It is a vector of real numbers between $0$ and $1$. The size of $D_i$ is equal to the total number of individuals at stage $i$. For an individual $j$, $D_i[j]$ is the  probability of being selected at $j$th stage, i.e., probability that $\hat{Y_i} = 1$. Assume $k$ stages in the entire process.

\subsubsection{$\bm{DF^2}$ Algorithm.} The $DF^2$ algorithm solves the following optimization problem at every stage $i \in \{1, 2, \dots k\}$:

\begin{equation}\label{prob-opt}
\begin{aligned}
\max_{D_i} \quad & P(Y = 1 | \hat{Y_i} = 1)\\
\textrm{s.t.} \quad & P(\hat{Y_i} = 1) = \alpha_i\\
  & f_i(\hat{Y_i}) = 0    \\
\end{aligned}
\end{equation}

\noindent where $P(Y = 1 | \hat{Y_i} = 1)$ is the precision of the decisions taken at stage $i$, $P(\hat{Y_i} = 1) = \alpha_i$ is the budget constraint at stage $i$, and $f_i(\hat{Y_i}) = 0$ is the fairness constraint at stage $i$.

\smallskip

\noindent Precision $P(Y = 1 | \hat{Y_i} = 1)$ can be replaced by an empirical estimate using the conditional risk scores $P(Y|X_1, \dots, X_i)$ of the individuals at stage $i$, where \{$X_1, \dots, X_i\}$ is the set of features observed by the algorithm at stage $i$. We assume that the sensitive attribute is observed in the first stage, i.e. $Z\in X_1$. Let $n_i$ be the total number of individuals at stage $i$,

$$P(Y = 1 | \hat{Y_i} = 1) \approx \frac{1}{\alpha_i} \sum_{j = 1}^{n_i} \frac{D_i[j]\cdot P_j(Y | X_1, \dots, X_i)}{n_i}$$

\noindent For budget constraint on the fraction of individuals selected at stage $i$, we define this fraction w.r.t to the total number of individuals in the first stage (i.e.  $n_1$). The constraint can be written as follows using an empirical estimate:

$$P(\hat{Y_i} = 1) \approx \frac{\sum_{j = 1}^{n_i} D_i[j]}{n_1}$$

\noindent Demographic parity fairness constraint is $P(\hat{Y_i} = 1 | Z = b) -  P(\hat{Y_i} = 1 | Z = w) = 0$; $Z$ being the sensitive attribute. Equality of opportunity fairness constraint is $P(\hat{Y_i} =  1| Y = 1, Z = b) - P(\hat{Y_i} =  1| Y = 1, Z = w) = 0$. The probabilities in fairness constraints can again be replaced by their empirical estimates and in terms of only the decision variables $D_i$, the sensitive attribute $Z$ and the conditional distribution $P(Y|X_1, \dots, X_i)$. We thus get a linear optimization problem in the decision variables $D_i$. Note that we only used the conditional distributions $P(Y|X_1 \dots X_i)$ in the optimization problem, and as discussed earlier, these distributions can be consistently estimated from incomplete data.  We enumerate the steps of $DF^2$ in a two stage process for more clarity in the supplementary material.

\subsubsection{Advantages of $\bm{DF^2}$.} While the primary advantage of $DF^2$ is that it doesn't require knowledge about the joint distributions $P(X_1, \dots, X_k)$ (hence the name \textit{detail-free}), it also offers another interesting advantage over the optimal algorithm of~\citet{emelianov2019price}. Theirs is essentially a centralized approach assuming that a single decision maker is controlling all stages of the selection process and thus, is able to find and enforce the optimal parameters for each stage. But in many realistic scenarios, independent agencies are responsible for implementing different stages without any communication or data sharing. The $DF^2$ algorithm is a decentralized approach and each stage can make its decisions independently of others, without communication.
\begin{figure*}
\begin{minipage}{0.5\textwidth}
\centering
\begin{subfigure}{.33\textwidth}
  \centering
  \includegraphics[width=\linewidth]{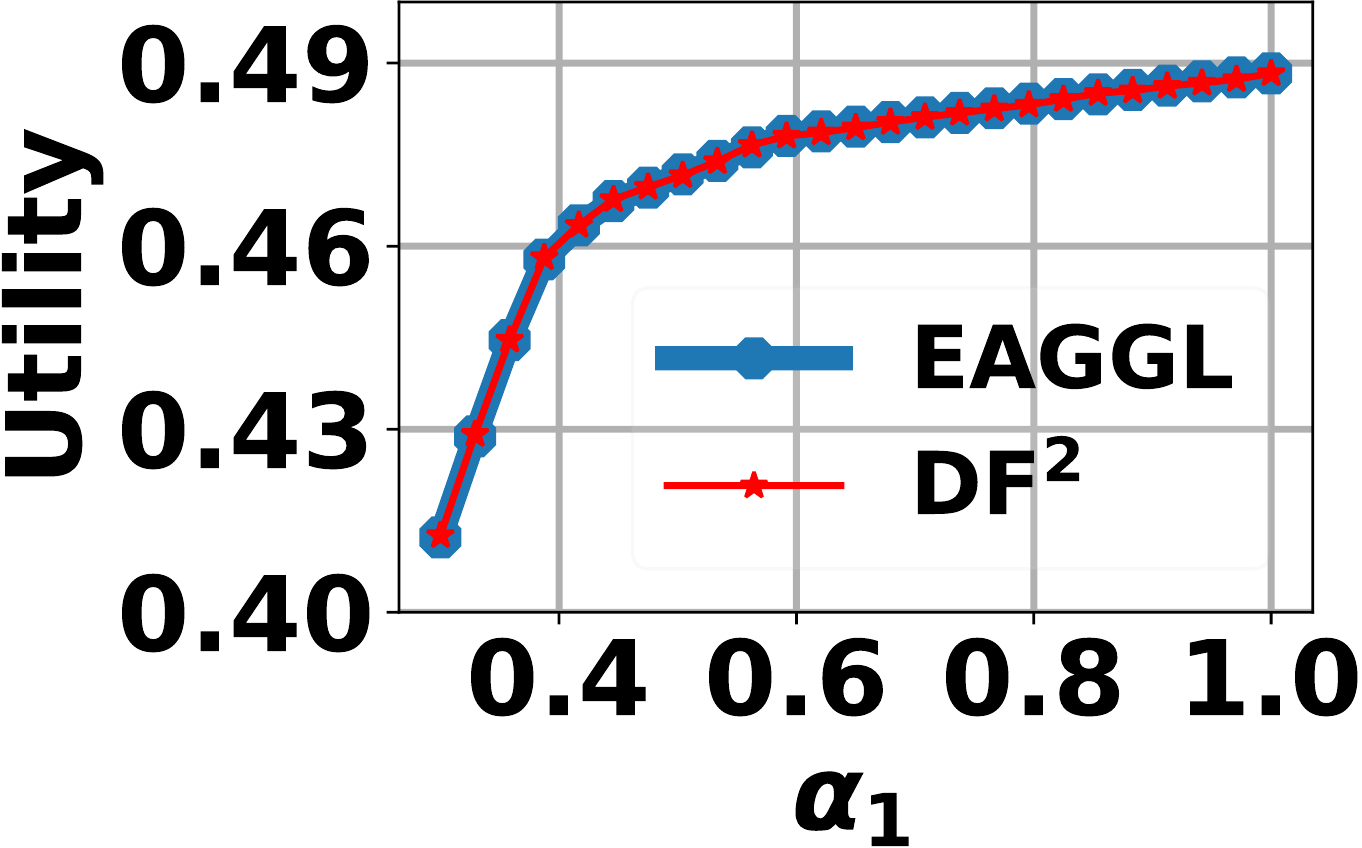}
  \caption{ADULT}
  \label{fig:adult_dp_utlity}
\end{subfigure}
\begin{subfigure}{.33\textwidth}
  \centering
  \includegraphics[width=\linewidth]{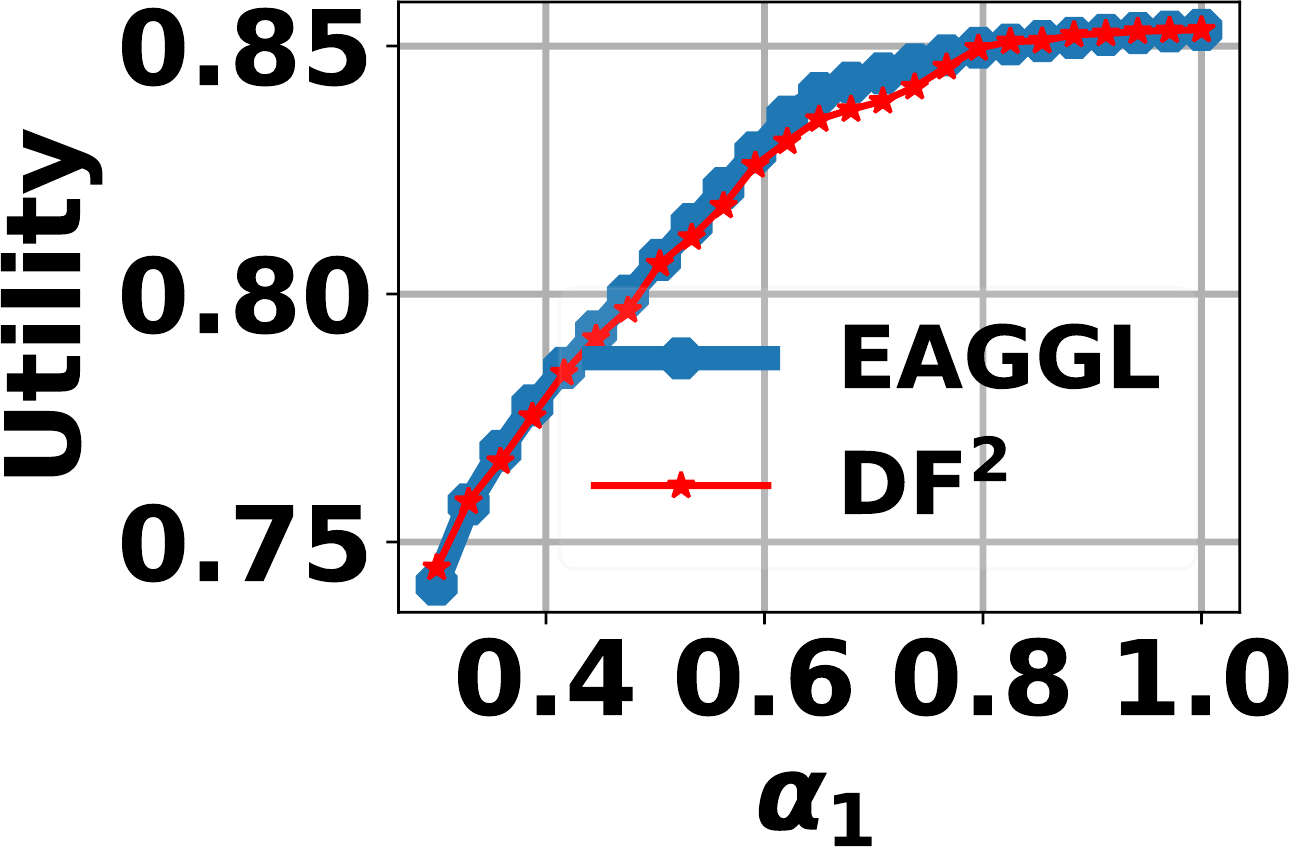}
  \caption{GERMAN}
  \label{fig:german_dp_utlity}
\end{subfigure}
\begin{subfigure}{.32\textwidth}
  \centering
  \includegraphics[width=\linewidth]{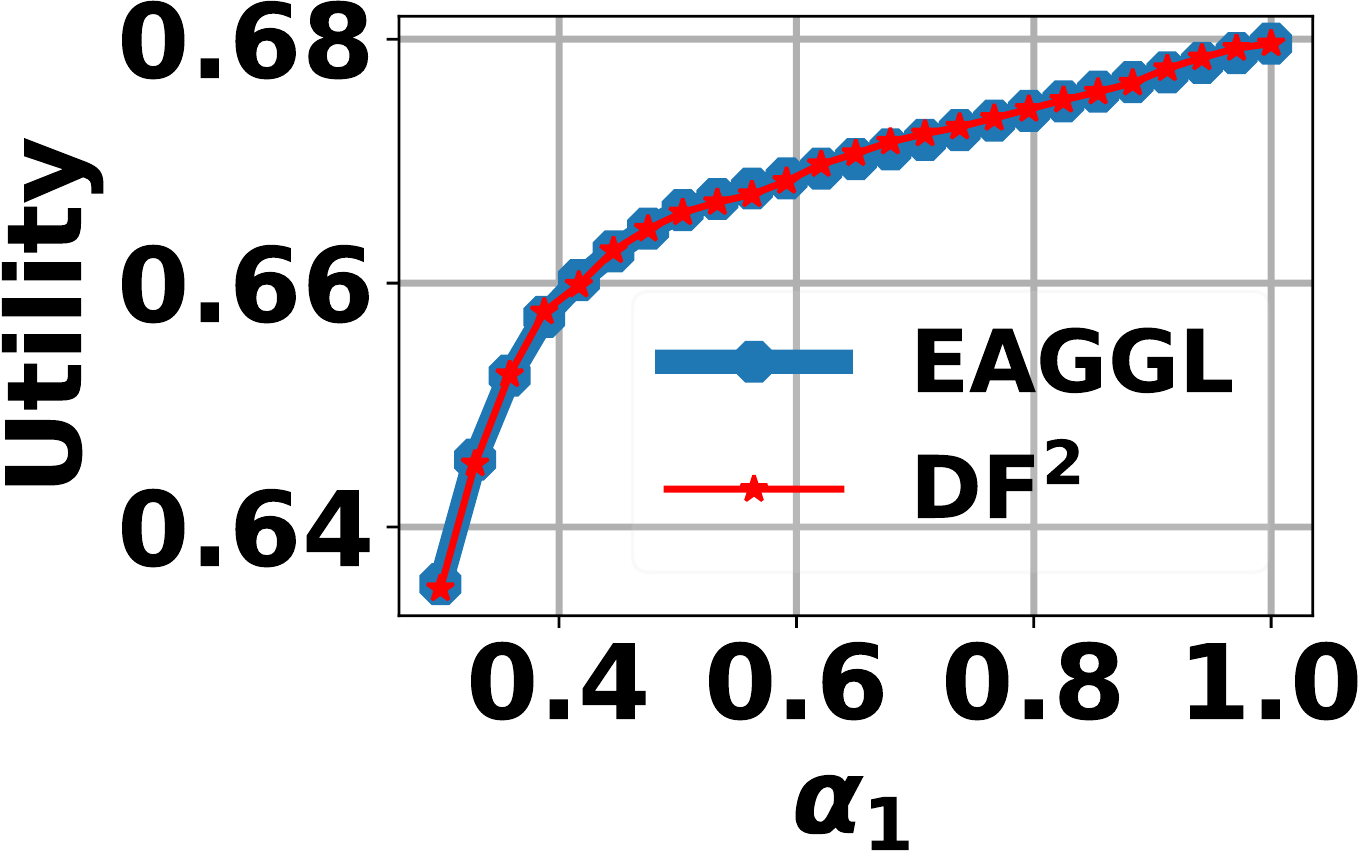}
  \caption{COMPAS}
  \label{fig:compas_dp_utlity}
\end{subfigure}
\caption{Demographic Parity Fairness Constraints}
\label{fig:dp}
\end{minipage}%
\begin{minipage}{0.5\textwidth}
\centering
\begin{subfigure}{.33\textwidth}
  \centering
  \includegraphics[width=\linewidth]{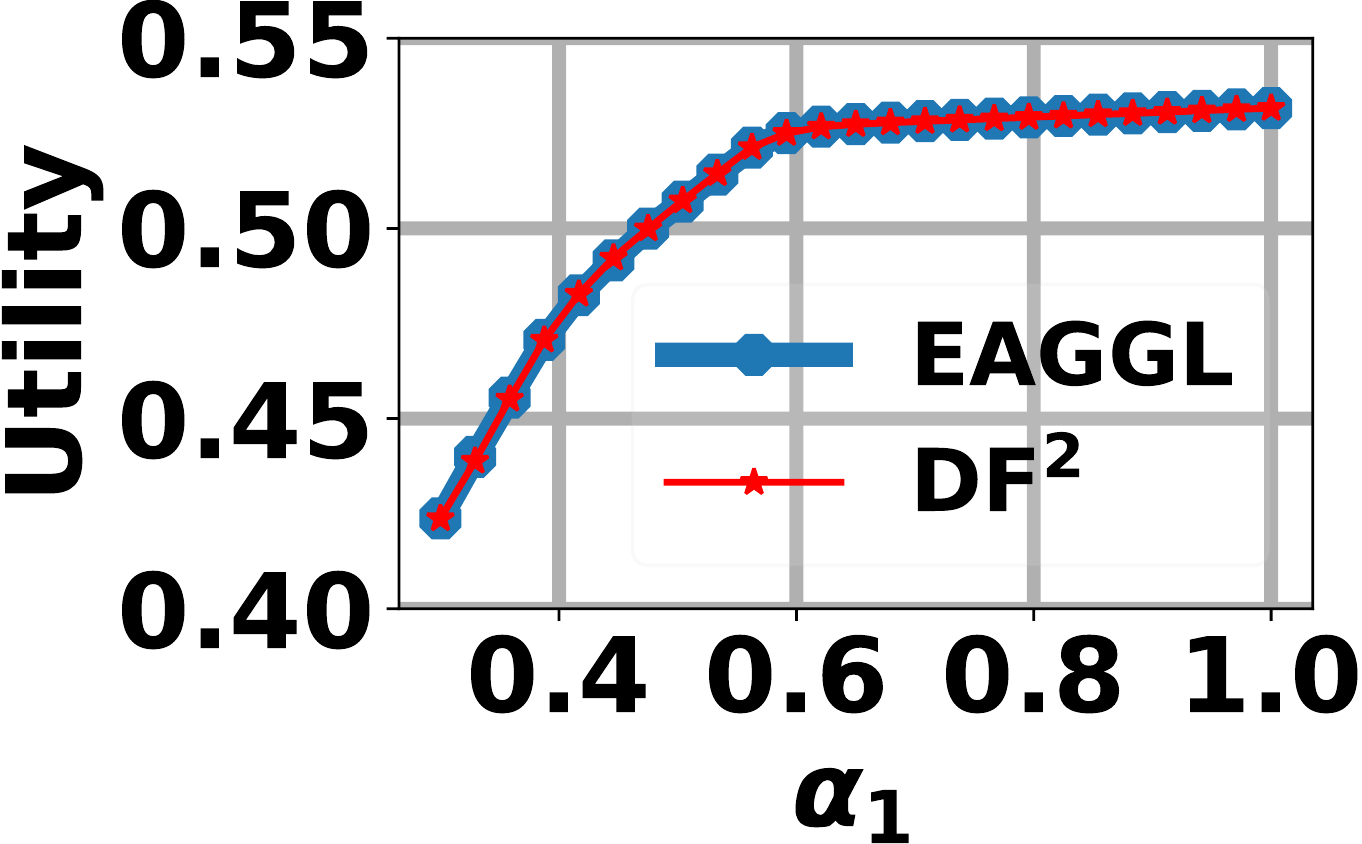}
  \caption{ADULT}
  \label{fig:adult_eo_utlity}
\end{subfigure}%
\begin{subfigure}{.33\textwidth}
  \centering
  \includegraphics[width=\linewidth]{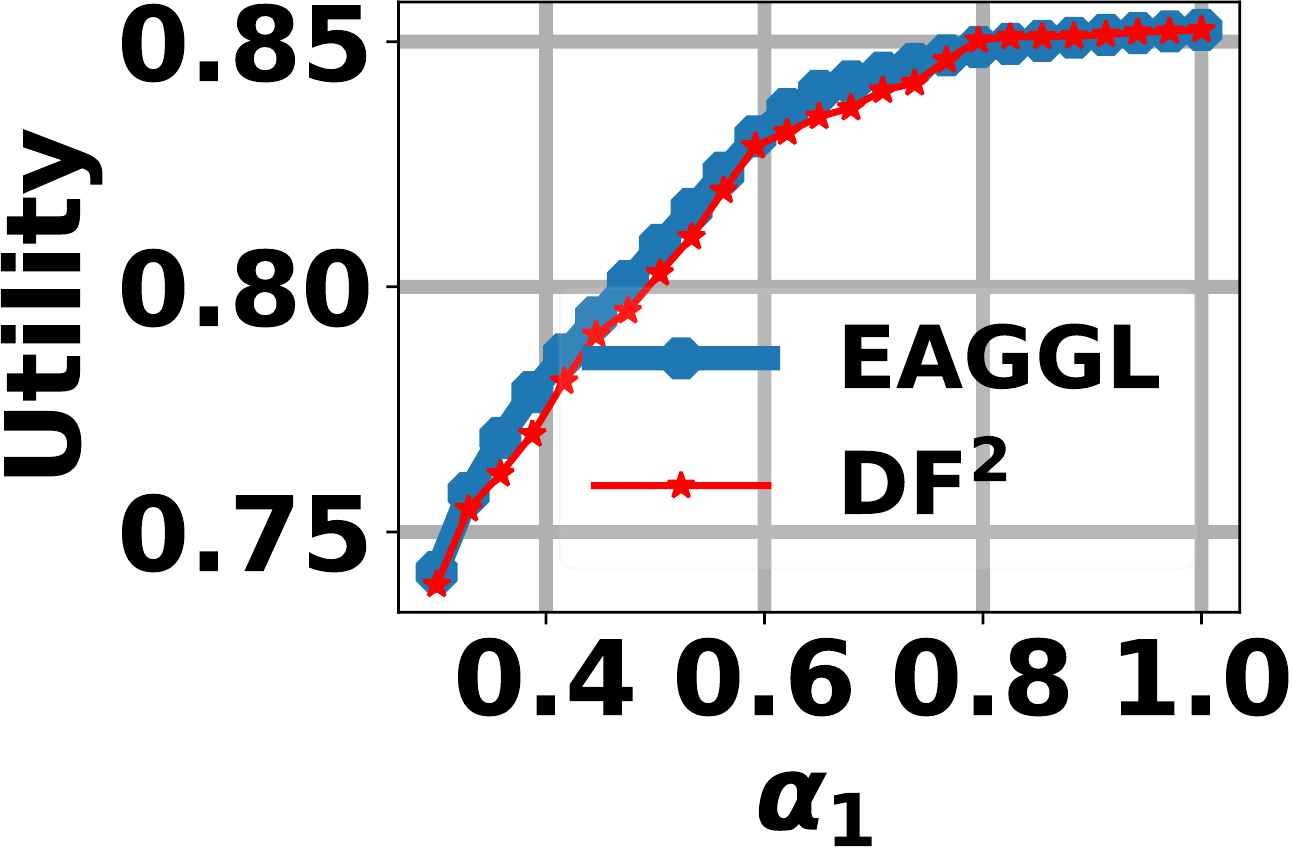}
  \caption{GERMAN}
  \label{fig:german_eo_utlity}
\end{subfigure}%
\begin{subfigure}{.32\textwidth}
  \centering
  \includegraphics[width=\linewidth]{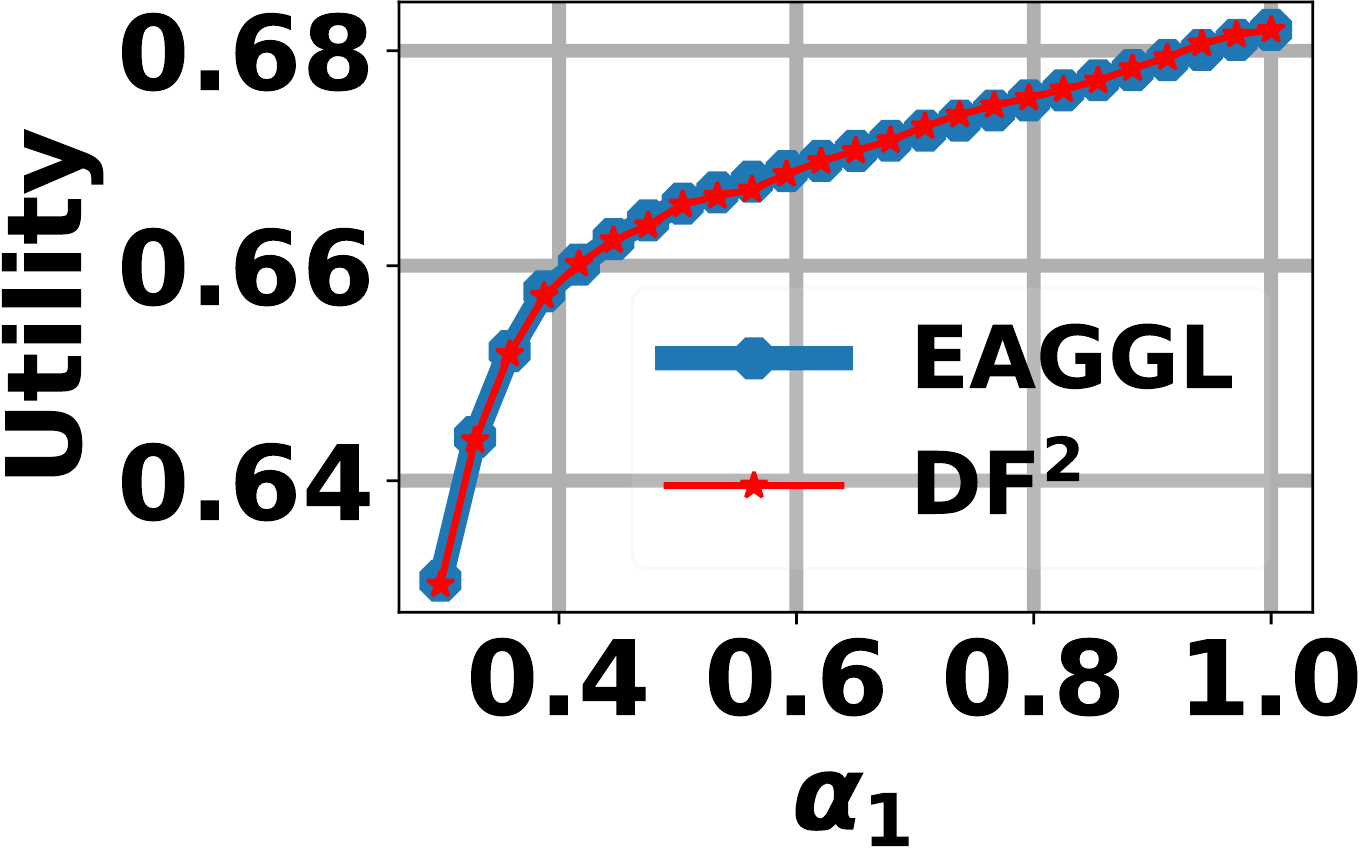}
  \caption{COMPAS}
  \label{fig:compas_eo_utlity}
\end{subfigure}%
\caption{Equality of Opportunity Fairness Constraints}
\label{fig:eo}
\end{minipage}
\end{figure*}

\subsection{Theoretical Analysis}
While our proposed algorithm offers the advantages of being detail-free and decentralized, an important question is whether it returns a solution with lower precision compared to an optimal algorithm (an oracle algorithm that knows the joint distribution)? Consider a two stage selection process with fairness constraints and $\alpha_1$ and $\alpha_2$ budget constraints in the two stages. Let $r_1$ and $r_2$ denote the random variables $P(Y|X_1)$ and $P(Y|X_1, X_2)$. We know from~\cite{menon2018cost, corbett2017algorithmic} that optimization problem of the form used in $DF^2$ results in subgroup specific thresholds on the risk scores (i.e. individuals above their subgroup specific threshold are selected and everyone else is rejected). Let $\delta_z(\alpha_1)$ be the threshold (for a subgroup $z$) on $r_1$, and let $\delta_z(\alpha_2)$ be the threshold on $r_2$ assuming that the first stage doesn't exist (i.e. all individuals from the first stage are made available to the second stage). Let $\alpha_2 < \alpha_1 < 1$, which implies that $\delta_z(\alpha_2) > \delta_z(\alpha_1)$ since lower budget means higher threshold. Further, let $D$ denote $DF^2$'s \textit{output decision vector} for all individuals that were initially present in the first stage (i.e. $D[j] = 1$ for individuals selected in the final stage by the $DF^2$ algorithm and $D[j] = 0$ for everyone else). Let $D^
O$ be the output decision vector of an oracle algorithm for same pool of individuals, budget and fairness constraints. We further assume:

\begin{assumption}[Coherent Features Assumption]
For any $m', m \in [0,1]$ such that $m' \geq m$, the following is true
$$P(r_1 \geq \delta_z(\alpha_1) | r_2 \geq m') \geq P(r_1 \geq \delta_z(\alpha_1) | r_2 \geq m)$$
\end{assumption}

\smallskip

\noindent This fairly weak assumption intuitively means that knowing that someone has a higher $r_2$ does not decrease the probability of their $r_1$ being above a threshold. The assumption is quite natural if the feature sets $X_1$ and $X_1, X_2$, in expectation, provide coherent information about the outcome $Y$. For example, when filtering job candidates, extra information (such as letters of recommendation) may be collected at a second stage. The assumption states that, on average, LoRs and first-stage scores don't provide conflicting information.

\begin{theorem}
Under the coherent features assumption, for budget constraints given by $\alpha_{1}$ and $\alpha_{2}$ on the fraction of individuals selected in the first and second stages, respectively, and demographic parity or equal opportunity constraint w.r.t outcome $Y$, the probability that $DF^2$'s output decision vector $D$ has lower precision compared to the oracle's output decision vector $D^O$ (i.e. $P(D \neq D^O)$) is upper bounded as
\[
P(D^O \neq D) \leq P(r_1 < \delta_z(\alpha_1) | r_2 = \delta_z(\alpha_2)),
\]
where $r_1$ and $r_2$ denote the random variables $P(Y|X_1)$ and $P(Y|X_1, X_2)$, $\delta_z(\cdot)$ denotes the subgroup specific threshold discussed above, and $P(\cdot)$ denotes probability.
\end{theorem}

\smallskip

\noindent The above theorem shows that the probability of getting a suboptimal solution in $DF^2$ depends on: 1) the distribution of the risk scores $r_1$ and $r_2$ (which are derived from the feature sets $X_1$ and $X_1, X_2$ respectively) used in the two stages, and 2) the budgets $\alpha_1$ and $\alpha_2$ of the two stages. With increasing positive correlation between $r_1$ and $r_2$ (e.g., more agreement between first-stage scores and LoR), and an increasing gap between the budgets $\alpha_1$ and $\alpha_2$ (e.g. final hiring being very selective compared to first stage screening), the probability of getting a suboptimal solution decreases. The analysis can be similarly extended to a $k$-stage process. In the next section, we empirically show that on real datasets, our algorithm's performance is similar to the oracle algorithm.

\subsection{Empirical Analysis}\label{sec:df2-results}
\subsubsection{Datasets.} We use three real datasets for evaluation that were also used by~\citet{emelianov2019price}. In the ADULT dataset~\cite{dua2017uci}, the outcome variable is \textit{salary} and the sensitive attribute is \textit{gender}. In the COMPAS dataset~\cite{compas-dataset}, the outcome variable is \textit{recidivism} within two years and sensitive attribute is \textit{race}. In the GERMAN dataset~\cite{dua2017uci}, the outcome variable is \textit{returns} which represents if an applicant paid back a loan and sensitive attribute is \textit{gender}. We discretize all the features using the same procedure as~\citet{emelianov2019price} (using their open source code on github). Details about the datasets and data processing are in the supplement.

\subsubsection{Methodology} We simulated a two stage process using the same sequence of feature observations in the two stages as ~\citet{emelianov2019price}.  In the ADULT dataset, the first stage observes \textit{gender, age, education} while the second stage includes the previous features plus \textit{relationship} and \textit{native country}. In the COMPAS dataset, the first stage observes \textit{race, young (younger than 25) and drugs (arrest due to selling or possessing of drugs)} while the second stage includes the previous features plus \textit{old (older than 45), gender, long sentence (sentence was longer than 30 days)}. In the GERMAN datatset, the first stage observes \textit{gender, job (is employed), housing (owns house)}, while the second stage includes the previous features plus \textit{savings (more than 500 DM), credit history (all credits payed back duly) and age (older than 50)}. We refer the method by~\citet{emelianov2019price} as EAGGL and consider it the \emph{oracle} method since it uses a centralized algorithm assuming knowledge of the full joint distribution. We compare the utility (equivalent to precision as discussed earlier) achieved by the $DF^2$ and the EAGGL (oracle) algorithms under fairness constraints (both methods return solutions satisfying fairness constraints). The purpose of our empirical analysis is to establish that it is possible to design fair algorithms without using non-recoverable distributions, with similar performance. The algorithm was implemented was in Python using CVXPY.

\subsubsection{Observations.} Figures~\ref{fig:adult_dp_utlity}, \ref{fig:german_dp_utlity} and \ref{fig:compas_dp_utlity} show the comparison between utility achieved by DF$^2$ and that by EAGGL in the three datasets, with demographic parity fairness constraints. The budget of the second stage was fixed at 0.3 ($\alpha_2 = 0.3$) to match the simulation parameters used by~\citet{emelianov2019price}. The budget of the first stage $\alpha_1$ was varied from $0.3$ to $1$. Clearly, the utility of DF$^2$ matches the optimal utility as achieved by EAGGL. Figures~\ref{fig:adult_eo_utlity}, \ref{fig:german_eo_utlity} and \ref{fig:compas_eo_utlity} show the same trend for equality of opportunity fairness constraints as well. We note that EAGGL uses inequality in the budget constraints in non-final stages unlike $DF^2$ which uses equality in the budget constraints in all the stages. However, in our experimental data, we found that despite the inequality constraints, EAGGL uses the entire budget available to maximize the precision. So, this difference in optimization constraints is not a factor to be concerned about while drawing conclusions from the results. We observed similar results in a three stage process as well (details in the supplement).
\subsubsection{Global Fairness.} 
\begin{figure}[ht]
\centering
\begin{subfigure}{.33\columnwidth}
  \centering
  \includegraphics[width=\linewidth]{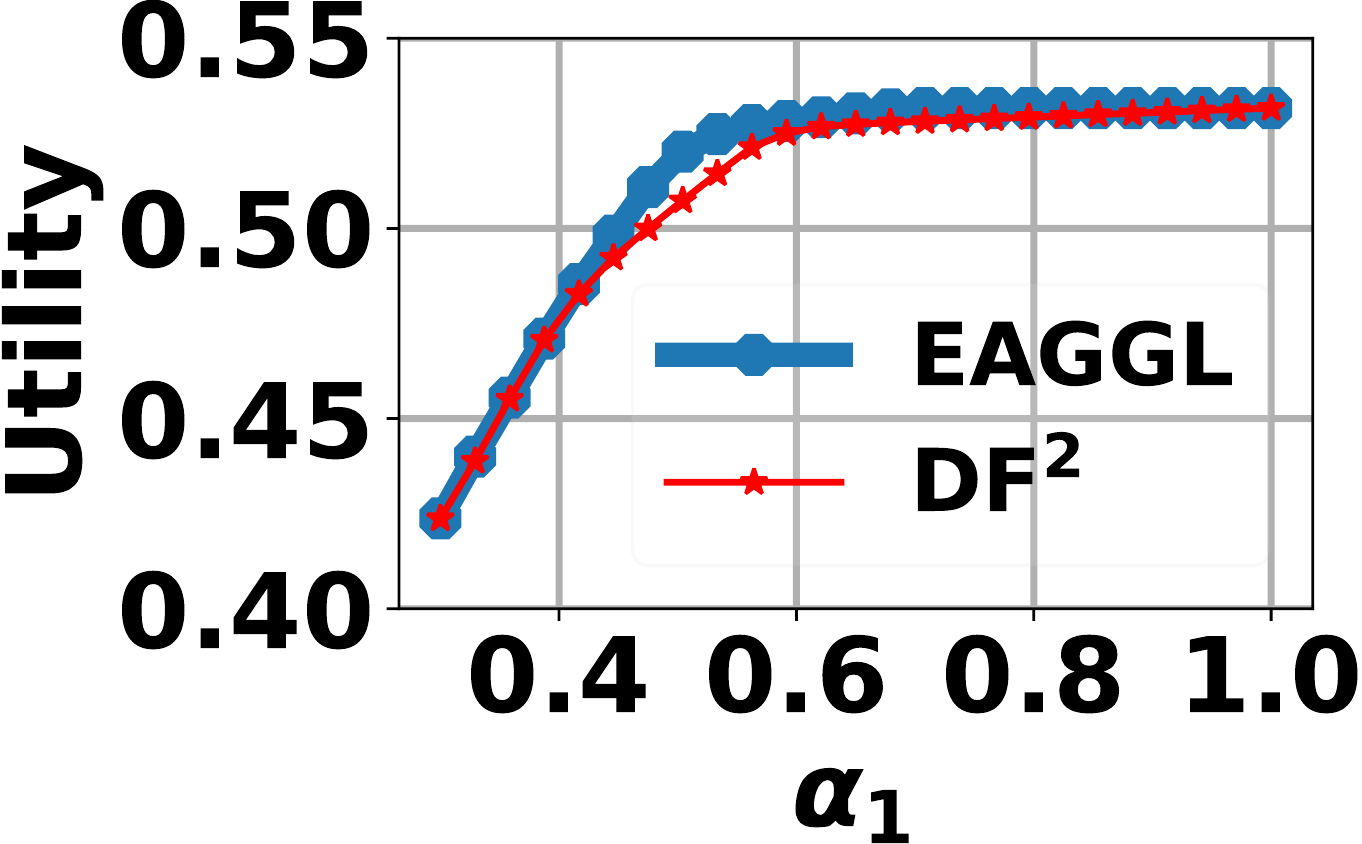}
  \caption{ADULT}
  \label{fig:adult_2def_utlity}
\end{subfigure}%
\begin{subfigure}{.33\columnwidth}
  \centering
  \includegraphics[width=\linewidth]{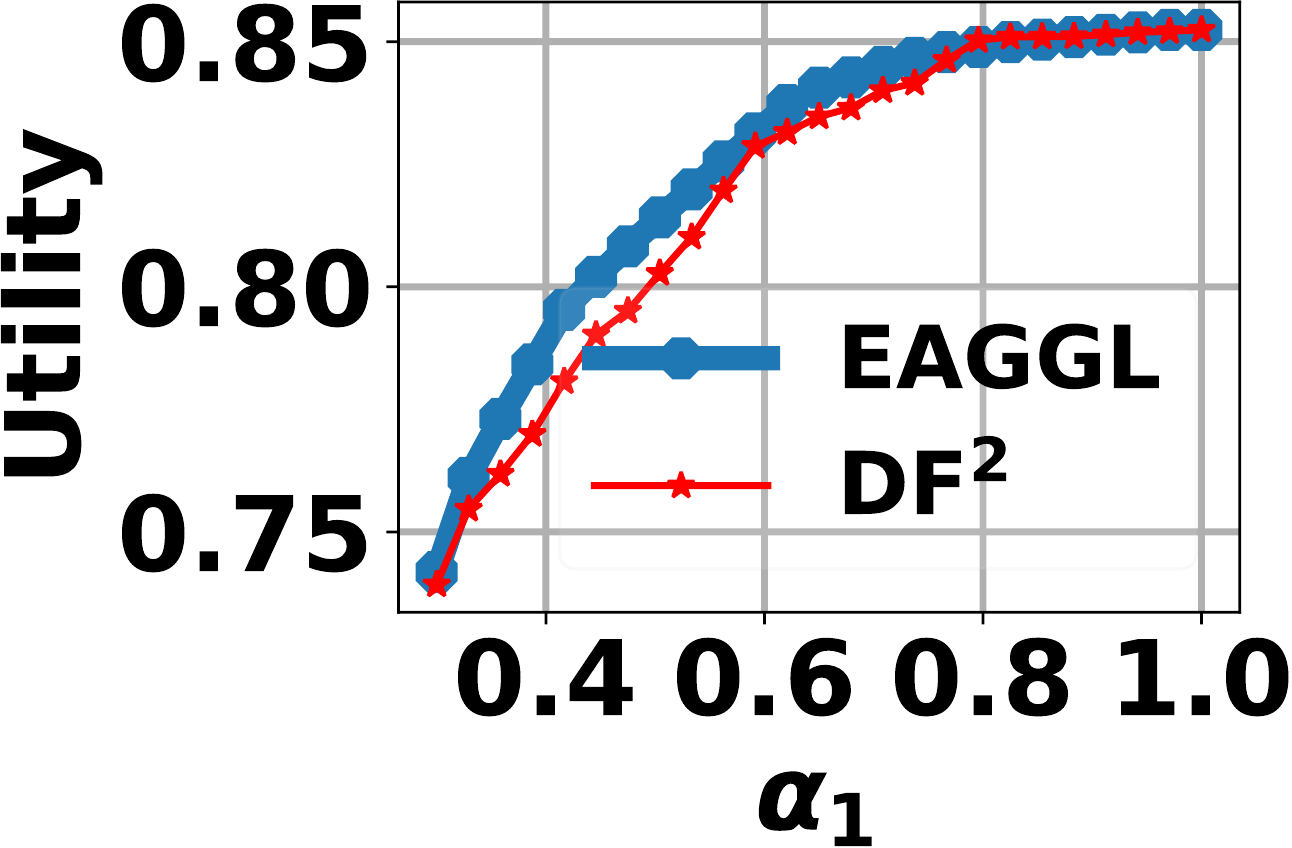}
  \caption{GERMAN}
  \label{fig:compas_2def_utlity}
\end{subfigure}%
\begin{subfigure}{.33\columnwidth}
  \centering
  \includegraphics[width=\linewidth]{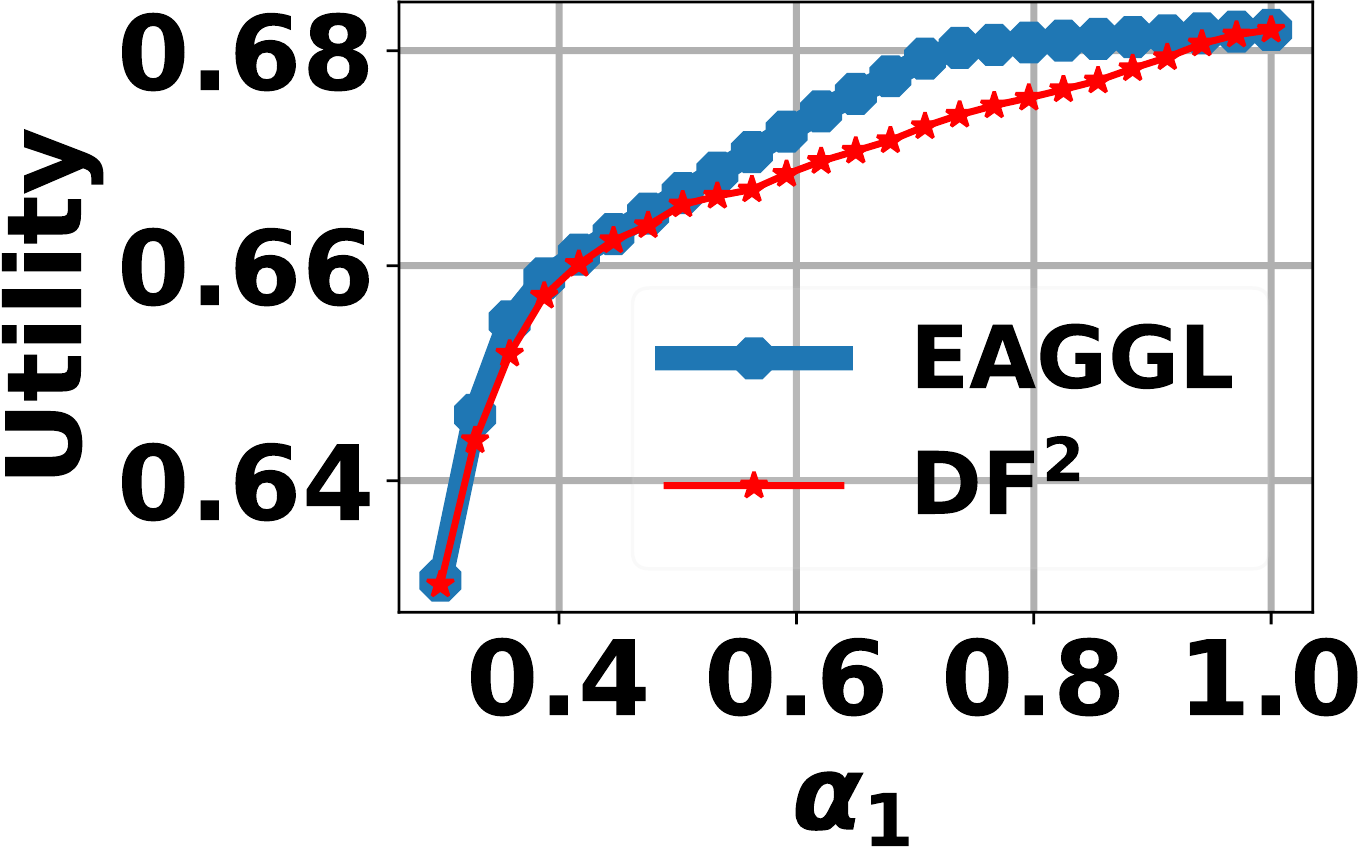}
  \caption{COMPAS}
  \label{fig:german_2def_utlity}
\end{subfigure}%
\caption{EAGGL(Only Global Fairness) vs $DF^2$}
\label{fig:all_2def}
\end{figure}

Fairness constraints we discussed so far are also referred to as ``local" in multi-stage settings~\cite{emelianov2019price}. Unlike local fairness (that requires fairness at every stage), the concept of global fairness requires only the final stage to enforce fairness. Local fairness is a stronger concept (and in theory, a more costly one) compared to global fairness. As shown in~\cite{emelianov2019price, bower2017fair}, \emph{local fairness implies global fairness but the opposite is not true}. Local fairness may be a more desired property in decentralized settings where individual stages are accountable for their own decisions and there is no central authority accountable on the behalf of all of them. $DF^2$, by design, satisfies both local and global fairness. Figure~\ref{fig:all_2def} shows that the difference in utilities of $DF^2$ with local (and thus also global) fairness constraints and EAGGL with only global fairness constraints is marginal.

\section{Conclusions}
Our results show that data missingness has significant implications for fairness algorithms and that it is important to model the missingness mechanism to accurately train fair classifiers. We used causal graphs to model missingness mechanisms in data for various real-world scenarios and discussed implications of missingness on correctly estimating distributions used in common fairness algorithms. Our results also provide useful perspective on correct data collection practices for fairness in machine learning. While the graph structures we discussed are not exhaustive, our framework can be applied to new settings by utilizing the causal structure to determine recoverability of quantities of interest. As an example, we applied our framework to develop a fair multi-stage decision making algorithm that requires no centralization between stages and uses only distributions that are recoverable from incomplete data.

\section*{Ethical/Societal Impact Statement}
This paper discusses a fundamental assumption of availability of uncensored training data behind many fairness algorithms, that mostly does not hold in practice. We hope that the paper will be helpful in designing fair algorithms in the absence of this assumption. While disposing of this assumption may be possible in many settings, there may be new assumptions that may have to be introduced in the process. Specifically, we assumed knowledge of the causal graph and in the case of multi-stage decision making, the coherent feature assumption. It is always important to evaluate the suitability of such assumptions in specific applications since the performance and fairness of the algorithm may be unpredictable if the assumptions are not suitable for the given application.

\bibliography{bib-aaai21}

\end{document}